\documentclass[10pt]{article}
\usepackage[margin=0.8in]{geometry}
\usepackage[T1]{fontenc}
\usepackage{lmodern,microtype,amsmath,amssymb,booktabs,array,longtable}
\usepackage[colorlinks=true,allcolors=blue]{hyperref}
\usepackage{enumitem}
\setlist{nosep}
\usepackage{indentfirst}
\usepackage{float}
\title{\Large Expert L2 English Speech Evaluation on a Gaming PC:\\[3pt]\large Accuracy, Fluency, and Prosody with XLS-R, XEUS, and PhoneticXeus}
\author{Richard Rose \enspace\textbar\enspace HUFS \enspace\textbar\enspace \texttt{richard.rose@hufs.ac.kr}}
\date{}
\usepackage{pdflscape}
\begin{document}
\maketitle
\begin{abstract}
\textit{oollim} is a locally executable system for evaluating L2 English pronunciation accuracy, fluency, and prosody on a consumer computer. The system combines pretrained XLS-R, XEUS, and PhoneticXeus representations with reference-text features and ridge regressions trained on SpeechOcean762, the Korean Learners' L2 Speech Corpus (English), J-AESOP, and their combinations. On 305 independently held-out recordings assessed by three expert raters, several ridge configurations substantially outperform archival Microsoft Azure Pronunciation Assessment API data in both Spearman correlation and quadratic weighted kappa across all three constructs. These results support expert speech evaluation on consumer hardware using pretrained representations and construct-specific regressions, without neural-model fine-tuning or calibration on the private evaluation data.
\end{abstract}

\section{Introduction}
Pronunciation evaluation requires distinct judgments of accuracy, fluency, and prosody. Accuracy concerns the realization of expected consonants and vowels \cite{deterding,cai}; fluency concerns speech continuity, including speed, pauses, and repairs \cite{kormos,bosker}; prosody concerns rhythm, stress, intonation, and phrasing \cite{cole,cutler}. These related dimensions inform pronunciation judgments but are not interchangeable \cite{cai,isaacs}. Segmental accuracy concerns distinctions between speech sounds, whereas intelligibility concerns what listeners understand and comprehensibility their perceived processing difficulty; a noticeable accent need not prevent understanding \cite{munro}. An accuracy score does not directly measure intelligibility. Fluency research distinguishes underlying processing efficiency, observable timing, and perceived smoothness \cite{segalowitz}. Speech rate, pause placement and duration, and repairs contribute differently to listener judgments, so fluency is not simply speaking quickly \cite{kormos,bosker}. Prosody organizes speech through lexical stress, prominence, phrasing, and intonation, with pitch, duration, and intensity providing interacting cues \cite{cole,cutler}. Appropriate stress can improve listeners' recall and evaluations of a speaker \cite{hahn}; contextual appropriateness therefore matters alongside acoustic variation. Although their acoustic correlates overlap, the three constructs require separate targets and useful measurements for educators; the scores need not come from the same representation combination or scoring function. XLS-R, XEUS, and PhoneticXeus provide complementary sources of speech information. This study examines whether these existing models, combined with modest supervised scoring functions, can support expert L2 English evaluation on an ordinary gaming PC. With computationally expensive pretraining supplied by the original model developers, the contribution is the selection and combination of their representations, reference-text evidence, and corpus-specific assessment supervision. The central evidence is an external evaluation on independently collected, privately held recordings, while training-corpus development results serve model selection and reproducibility and are provided in the repository. The external comparison includes Azure pronunciation assessment \cite{azure}, seven ridge-training families, and two human agreement baselines, with both rank association and numerical agreement measured. This design separates three questions: whether scores rank recordings similarly to human judgments, whether they agree on the numerical scale, and how that agreement compares with agreement among the human raters \cite{williamson}.

\section{Speech models and related work}
Automated assessment has long combined acoustic evidence with supervised mappings to human judgments. SpeechRater derives pronunciation, fluency, and other proficiency features from recognized speech and uses learned scoring models \cite{speechrater}. Neural assessment extends this approach by learning representations and their relationship to rated performance. GOPT uses phonetic posterior features and canonical phone identities in a transformer trained jointly on phone-, word-, and utterance-level human annotations \cite{gopt}. On SpeechOcean762, the 2,500-utterance training and test partitions provide the same assessment targets on different recordings \cite{speechocean,gopt}. The scoring model is trained on human ratings and evaluated against ratings of held-out recordings. Kim et al. adapt wav2vec 2.0 and HuBERT with CTC, then combine contextual representations and reference text in a bidirectional LSTM scoring model, evaluating against expert scores on separate partitions of Korean learner speech and SpeechOcean762 \cite{kim}. Cai et al. connect hierarchical neural assessment to applied-linguistics constructs, adapting scoring toward human judgments of segmental, suprasegmental, and temporal performance \cite{cai}. Saito et al. use regression over phonological, prosodic, and temporal features to predict comprehensibility, testing generalization on additional recordings and an interview task \cite{saito}. These studies train scoring models against human ratings and assess multiple dimensions or transfer to new data, although differences in targets, rating panels, and metrics prevent direct numerical comparison with \textit{oollim}. XLS-R extends cross-lingual self-supervised speech representation learning \cite{xlsr}; XEUS learns multilingual representations across thousands of languages \cite{xeus}; PhoneticXeus adds phone recognition and supplies hidden representations and phone evidence \cite{phonetic}. In \textit{oollim} their pretrained weights remain unchanged, while ridge regression learns regularized linear mappings to assessment labels \cite{ridge}. \textit{oollim} combines these representations with reference-phone diagnostics and construct-specific regressions trained on one or more corpora, then evaluates them on an independently collected passage-reading cohort to test transfer beyond holding out recordings within a training corpus. Earlier neural-head, temporal, and acoustic-feature experiments motivated the ridge baseline; the evaluation does not cover every neural assessment architecture. The comparison therefore focuses on the external transfer of compact scoring models.

\section{Method}
\subsection{Audio, reference text, and representations}
Whole-recording embeddings and phone diagnostics derived from required reference text provide the scoring inputs. FFmpeg \cite{ffmpeg} decodes audio-bearing containers; a common preprocessing path averages channels to mono and resamples to 16 kHz. The full recording is processed as one utterance, without averaging separately scored sentences. Each encoder adapter returns frame embeddings and valid-frame lengths or masks, with XLS-R 300M, XEUS, and PhoneticXeus each providing 1,024-dimensional frame representations in the configurations used here \cite{xlsr,xeus,phonetic}. Statistics pooling \cite{pooling}, using masked temporal means and population standard deviations, with $10^{-5}$ added to the variance before the square root, yields 2,048-dimensional utterance vectors. Model revisions and implementation checksums are pinned in the repository. A pinned CMU pronunciation dictionary supplies ARPAbet reference pronunciations \cite{cmudict}. ARPAbet--IPA correspondences are documented in CMU dictionary tools \cite{cmutools}; the versioned mapping in \textit{oollim} converts dictionary phones to the PhoneticXeus token inventory, including explicit diphthong splitting and unstressed-vowel handling. PhoneticXeus provides frame-level phone probabilities and a greedy connectionist temporal classification (CTC) sequence \cite{phonetic,ctc}. Seventeen diagnostics summarize phone probabilities and entropy \cite{shannon}, blank behavior, predicted and expected phone rates, reference coverage, relative sequence length, edit operations \cite{editdistance}, and CTC negative log-likelihoods normalized by frame and reference-token counts \cite{ctc}. The CTC calculation sums valid paths for the expected phone sequence rather than assigning a single forced alignment \cite{ctc}. These features combine pronunciation, timing, and uncertainty evidence; they are not error-free phonetic diagnoses. Out-of-vocabulary references are rejected explicitly. The evaluated accuracy configuration uses XEUS + PhoneticXeus + IPA diagnostics, whereas fluency and prosody use XLS-R + XEUS. The scoring pipeline can select a representation combination and training family separately for each construct.

\subsection{Training corpora and label mappings}
Three research corpora supply assessment labels mapped to a common 0--10 range. SpeechOcean762 supplies utterance accuracy, fluency, and prosodic ratings on the 0--10 scale. The official SpeechOcean762 training partition is divided by speaker into 2,000 training and 500 development utterances; the official test partition is excluded. The Korean Learners' L2 Speech Corpus (English) \cite{han} is distributed as AIHub dataset 71463 \cite{aihub}. Eligible Training PRN recordings yield 2,699 training and 636 development examples. The corpus rubric rates pronunciation accuracy (segment clarity and errors) and prosodic fluency (stress, rhythm, intonation, rate, and pauses) on 1--5 scales \cite{han}. The supplied articulation score trains accuracy, while the prosodic-fluency score trains both fluency and prosody after mapping with $2.5(y-1)$. J-AESOP \cite{aesop} supplies 180 Japanese L2 speakers reading the North Wind and Sun passage (task 6\_01). Deterministic speaker partitions allocate 108 recordings to training, 36 to development, and 36 to a reserved partition; native controls are excluded. Ratings from 16 raters across three sections are averaged per construct. Segment ratings supply accuracy, alongside fluency and prosody; $10(y-1)/9$ maps the 1--10 source scale to 0--10. Partitions are speaker-disjoint within each corpus to separate development from fitting \cite{cawley}. In the Korean Learners' L2 Speech Corpus (English), recorder identities linked by identical decoded audio are grouped before splitting; duration and dictionary-coverage checks determine eligibility. Exact sampling rules, audio audits, mappings, and development results are retained in the reproduction package. Equal numerical ranges do not establish score equivalence \cite{dorans}; the shared prosody/fluency target does not provide an independent fluency annotation. The mappings permit joint fitting while retaining these differences in supervision.

\subsection{Ridge fitting and family selection}
Seven training families compare single-corpus and mixed-corpus supervision. The single-corpus families are \texttt{speechocean}, \texttt{korean}, and \texttt{aesop}; \texttt{korean} abbreviates the Korean Learners' L2 Speech Corpus (English) in model identifiers. The pairs are \texttt{koreanocean}, \texttt{aesopocean}, and \texttt{koreanaesop}, while the three-corpus family is \texttt{aesopkoreanocean}. In the matched family procedure, a recording from corpus $c$ receives weight $w_i=1/(K n_c)$, where $K$ is the number of included corpora and $n_c$ their respective training counts, giving each corpus equal total weight regardless of size. Training-only weighted means and standard deviations standardize features \cite{esl}; each feature block (one encoder vector or the IPA diagnostics) of width $d_b$ receives gain $\sqrt{2048/d_b}$ to balance contributions from blocks of different sizes. Coordinates with standard deviation below $10^{-8}$ use scale 1. Each construct fits a weighted ridge objective \cite{ridge,esl}
\begin{equation}
\min_{\beta,b}\quad\sum_i w_i(y_i-z_i^\top\beta-b)^2+\alpha\lVert\beta\rVert_2^2,
\end{equation}
where $z_i$ is the standardized, block-scaled feature vector, $y_i$ the target, $\beta$ the coefficients, $b$ the unpenalized intercept, and $\alpha$ the regularization strength. Four approaches share corpus partitions, training-only scaling, and separate construct coefficients. The \emph{1-grid} is $\{0.001,0.01,0.1,1,10,100\}$; the \emph{5-grid} is $\{0.00005,0.0005,0.005,0.05,0.5,5\}$. The names identify the grid and selection criterion: 1G-S is \emph{1-grid construct-wise correlation}, 1G-AMAE is \emph{1-grid pronunciation-accuracy error}, and 5G-AMAE is \emph{5-grid pronunciation-accuracy error}. Correlation means construct-specific development Spearman; error means development pronunciation-accuracy mean absolute error (MAE). Development metrics are averaged equally over included corpora, keeping tuning separate from external evaluation \cite{cawley}. Under accuracy-MAE selection, fluency and prosody share the penalty selected using accuracy labels for their common XLS-R + XEUS features, but retain separate scoring coefficients. Three-ratio banded ridge (3RB) instead permits different penalties for XEUS and XLS-R \cite{banded}: $\alpha_{\rm XLS-R}/\alpha_{\rm XEUS}\in\{1/4,1,4\}$, selected by construct-wise correlation (3RB-C) or pronunciation-accuracy error (3RB-A). The scalar search and equal-penalty fallback are specified in the supplement; 3RB accuracy uses the corresponding refined scalar-ridge head, without banding. Seven families therefore yield 35 configurations. SpeechOcean is freshly refitted under each condition; the earlier deployed version remains separately archived. The representation sensitivity analysis examined 315 family--construct--representation configurations on public development data and retained the overall selections under its prespecified 0.01 Spearman improvement threshold and corpus-degradation checks. Those representation selections differ from the descriptive external leaders below. No private evaluation ratings enter coefficient estimation, preprocessing fitting, hyperparameter selection, or calibration.

\subsection{Local execution and capacity}
Local operation was verified on consumer hardware using CPU inference. On the Windows gaming PC (Intel Core i7-11700K, 32 GB physical RAM, Ubuntu under WSL2 with approximately 24 GiB available RAM and 8 GiB swap), an archived 120-second capacity test completed in approximately 169 seconds, reaching 22.52 GiB peak process resident memory and 5.51 GiB peak system-wide swap use. The installed RTX 3080 was not required for that inference test. A separate Apple Silicon installation test processed an 8.21-second public recording in 42.13 seconds with four CPU threads, including decoding, all three encoders, phone features, and scoring. Software versions for the same WSL installation are recorded in the appendices. These single-run checks establish operation rather than representative throughput; the released duration limit is 120 seconds, and long recordings can require substantial RAM and swap.

\section{External evaluation}
\subsection{Independent recordings and ratings}
The collection comprised university EFL learners from China ($n=98$), Korea ($n=101$), and Japan ($n=107$), who recorded the same ``Please Call Stella'' passage on personal smartphones. One recording was excluded because of its length, leaving 305 samples. The three university professors had approximately 20--30 years of teaching experience, including pronunciation instruction, Cambridge English speaking assessment, and high-stakes speech rating. Based in those countries, the raters calibrated their judgments on separate recordings spanning proficiency levels, discussing discrepancies to establish shared criteria for accuracy, fluency, and prosody. Each professor then rated every recording independently in randomized order, blinded to participant identity and automated scores, using a 0--100 scale in five-point increments. The private evaluation recordings and ratings were held out from model training, hyperparameter tuning, and score calibration. The original human ratings and Azure API scores are archival; the seven ridge families were evaluated for the present study against those existing ratings. Azure scores were obtained in January 2026 using the \texttt{en-US} locale, with the Stella reference text supplied and each complete recording assessed as a whole. The returned accuracy, fluency, and prosody scores were retained. The service revision and remaining request options were not retained in the supplied evaluation archive; the comparison therefore concerns those recorded API outputs rather than a newly standardized Azure benchmark.

\subsection{Metrics and human baselines}
Evaluation distinguishes rank consistency from numerical agreement \cite{stemler,williamson}. Spearman's $\rho$ measures rank association \cite{spearman}. Quadratic weighted kappa (QWK) measures chance-corrected agreement with squared category-distance penalties \cite{cohen}; this study uses fixed integer categories 0--100 and rounding to the nearest integer, with exact half-points rounded upward. Model outputs are multiplied by ten; no calibration is fitted. Predictions are clipped to 0--100 for QWK, including bootstrap replicates, whereas Spearman retains raw predictions. A reference mean requires all three human scores; partial means are not substituted. Human--Human pairwise agreement averages the three pairwise statistics, whereas Human--Human consensus agreement averages each rater's agreement with the mean of the other two, never including that rater in their own reference. For each system, the three-rater row compares the system's scores with the mean of all humans, while the two-rater row averages the three separate agreement statistics against the three possible two-human means. Statistics are averaged, not observations or reference means pooled. The two-rater system row and Human--Human consensus row therefore provide the appropriate reference-size comparison when recomputed on the same recordings. Percentile 95\% intervals resample speaker clusters jointly, recomputing all component statistics within each replicate \cite{efron,field}. The supplementary marginal intervals retain the supplied evaluation export (2,000 replicates, seed 42). Additional paired comparisons use 1,200,000 bootstrap replicates per seed (42 and 2026), subtracting statistics on the same resampled recordings. Each unique recording URL is treated as one speaker. Bonferroni-adjusted intervals cover 510 contrasts: 35 configurations versus human consensus and Azure, Azure versus human consensus, and each of the 14 banded configurations versus the previous 1G-S/1G-AMAE/5G-AMAE leader for each construct and metric. The previous leaders are fixed from the preceding evaluation, rather than reselected within bootstrap samples. Seed 42 supplies the reported intervals; the second seed checks interval-exclusion stability. These post-evaluation comparisons condition on the observed raters and fitted models; adjustment covers the stated contrasts, not the preceding model search. The supplement gives full comparisons on the same 305 recordings, and the aggregate release retains three-rater and human-pairwise results.

\section{Results}
Table~\ref{tab:external} reports observed leaders and ties within .001. 3RB-C aesopocean leads both fluency metrics, while 3RB-A koreanaesop leads prosody correlation; the accuracy and prosody-QWK leaders remain unchanged. All displayed correlation leaders significantly exceed both individual-human and archival Azure baselines on matched two-rater references after correction across 510 contrasts. 1G-S-koreanocean achieves significantly higher accuracy QWK than the human baseline (0.821 versus 0.704; adjusted difference interval [0.048, 0.191]); 3RB-C-koreanocean closely matches that result. 3RB-C-aesopocean has slightly higher fluency QWK than the human baseline (0.751 versus 0.739), but the difference is not significant (adjusted interval [$-$0.055, 0.082]). 5G-AMAE-koreanocean's prosody QWK remains significantly below the human baseline. Gains over the previous fluency correlation, fluency QWK, and prosody correlation leaders are about 0.006--0.007, and none is statistically established after correction. Across all 35 configurations, 99 of 105 correlation advantages and 95 of 105 QWK advantages over Azure retain positive adjusted intervals in both seeds. Seven of 510 decisions are seed-dependent and reported as borderline; none changes the foregoing conclusions. External leaders remain descriptive selections.

\begin{table}[H]
\centering
\small
\setlength{\tabcolsep}{5pt}
\renewcommand{\arraystretch}{1.0}
\caption{Construct-specific leaders and ties ($N=305$).}
\label{tab:external}
\begin{tabular}{@{}lllcccc@{}}
\toprule
Construct & Family & Rule & $\rho$ & 95\% CI & QWK & 95\% CI\\
\midrule
Accuracy & \textbf{Human consensus} & -- & \textbf{0.809} & [0.769, 0.842] & \textbf{0.704} & [0.663, 0.743]\\
 & Azure & -- & 0.761 & [0.709, 0.804] & 0.161 & [0.129, 0.196]\\
 & aesopocean & 1G-S & \textbf{0.872} & [0.842, 0.894] & 0.598 & [0.563, 0.631]\\
 & aesopocean & 5G-AMAE & \textbf{0.872} & [0.841, 0.894] & 0.607 & [0.572, 0.640]\\
 & aesopocean & 3RB-C & \textbf{0.871} & [0.841, 0.894] & 0.596 & [0.561, 0.629]\\
 & aesopocean & 3RB-A & \textbf{0.871} & [0.840, 0.894] & 0.612 & [0.576, 0.645]\\
 & koreanocean & 1G-S & 0.853 & [0.816, 0.882] & \textbf{0.821} & [0.789, 0.847]\\
 & koreanocean & 3RB-C & 0.852 & [0.813, 0.881] & \textbf{0.821} & [0.789, 0.847]\\
\midrule
Fluency & \textbf{Human consensus} & -- & \textbf{0.807} & [0.768, 0.840] & \textbf{0.739} & [0.695, 0.777]\\
 & Azure & -- & 0.633 & [0.556, 0.698] & 0.078 & [0.056, 0.107]\\
 & aesopocean & 3RB-C & \textbf{0.883} & [0.854, 0.906] & \textbf{0.751} & [0.718, 0.779]\\
\midrule
Prosody & \textbf{Human consensus} & -- & \textbf{0.807} & [0.768, 0.842] & \textbf{0.742} & [0.701, 0.778]\\
 & Azure & -- & 0.737 & [0.672, 0.792] & 0.370 & [0.318, 0.421]\\
 & koreanaesop & 3RB-A & \textbf{0.889} & [0.860, 0.911] & 0.446 & [0.409, 0.479]\\
 & koreanocean & 5G-AMAE & 0.840 & [0.802, 0.872] & \textbf{0.675} & [0.632, 0.712]\\
\bottomrule
\end{tabular}
\par\smallskip
\begin{minipage}{\textwidth}
\footnotesize
\textit{Note.} Statistics average the three two-human references; humans are compared with the other two. 1G-S/1G-AMAE: 1-grid construct-wise correlation/pronunciation-accuracy error; 5G-AMAE: 5-grid pronunciation-accuracy error; 3RB-C/3RB-A: three-ratio banded with the respective criteria. Correlation is Spearman; error is MAE. 3RB accuracy uses the refined scalar head. Every configuration within .001 of either unrounded maximum among 35 combinations is shown; bold marks qualifying metrics and human baselines. Leaders are descriptive external selections. CIs are marginal 95\% bootstrap intervals (2,000 draws, seed 42); paired tests separately adjust 510 contrasts. No displayed leader required clipping.
\end{minipage}
\end{table}

\section{Discussion}
Scorer selection depends on construct and intended use: public-development selection prioritizes rank correlation, whereas external QWK favors koreanocean for accuracy and prosody and banded aesopocean for fluency. These external preferences are descriptive, not independently validated deployment choices. Azure scores average 23.63, 27.88, and 10.57 points above the three-human means for accuracy, fluency, and prosody, respectively; these offsets help explain low QWK despite positive rank associations. No correction was fitted to remove those offsets. XEUS alone approaches the retained combinations on public-development macro correlation (gaps of 0.0017 for accuracy, 0.0016 for fluency, and 0.0041 for prosody in the selected families), so the sweep does not establish a substantial benefit from every additional encoder. A lighter XEUS-only system requires separate external evaluation. Korean-only fluency and prosody share features and targets, so their predictions do not constitute independent learning of two constructs. The human comparison distinguishes the constructs: accuracy QWK exceeds the individual-human baseline with statistical support, fluency QWK is comparable in the observed results, and prosody QWK remains lower. These comparisons measure agreement with the same two-rater references, rather than agreement with an error-free ground truth. Agreement with human ratings supports a particular score interpretation, rather than every educational use of the scores \cite{kane}. The software, trained scoring weights, model configurations, and reproduction materials are available at \url{https://github.com/cburst/oollim}, subject to the licenses documented in the repository.

\section{Limitations}
The evaluation supports conclusions about one passage and three university EFL settings. Recording-level scores permit paired analysis, but URL-based grouping assumes one recording per speaker. The cohort excludes one long recording, and the 120-second operational limit constrains input coverage. Score-scale mappings do not establish equivalent constructs: the Korean Learners' L2 Speech Corpus (English) supplies shared prosody/fluency supervision, so differences among training families cannot be attributed solely to additional data. Speaker-disjoint fitting also cannot resolve possible upstream exposure of PhoneticXeus to learner corpora. The Azure service revision and remaining request options are unavailable, limiting exact replication of that comparator. The public-development ablation used a recorded selection rule, but neither bootstrap intervals nor contrast adjustment account for the entire model-search process. Broader claims require additional passages, populations, and intended uses \cite{kane}.

\section{Conclusion}
XLS-R, XEUS, and PhoneticXeus with construct-specific ridge scoring support local evaluation of L2 English accuracy, fluency, and prosody on consumer hardware. The evaluated families provide strong rank association across all three constructs, with banded penalties extending the range of scoring choices. Different leaders emerge for numerical agreement, supporting the reporting of both metrics.

\section{Generative AI Use Disclosure}
OpenAI Codex assisted with software development, experiment execution, and manuscript drafting and revision; the author is responsible for the submitted content and its verification.

\clearpage

\appendix
\section{Family and selection-rule comparisons}\label{sec:fulltables}\label{sec:selectionrules}
All panels use the same 305 recordings and two-rater references. Model statistics average agreement with the three possible two-human means; human consensus compares each human with the other two. Marginal intervals retain the 2,000-draw, seed-42 evaluation export. 1G-S and 1G-AMAE use the 1-grid $\{0.001,0.01,0.1,1,10,100\}$, selected respectively by construct-specific Spearman and accuracy MAE. 5G-AMAE uses accuracy MAE on the 5-grid $\{0.00005,0.0005,0.005,0.05,0.5,5\}$. The 5-grid converts the earlier summed-loss candidates $\{0.1,\allowbreak1,\allowbreak10,\allowbreak100,\allowbreak1000,\allowbreak10000\}$ by dividing by the original 2,000 SpeechOcean training utterances; the same fixed grid is used for every family. The conversion preserves penalty scale, rather than reproducing every earlier implementation detail. All fitting, scaling, and tuning use public data only.

Three-ratio banded ridge searches XLS-R/XEUS penalty ratios $r\in\{1/4,1,4\}$ for fluency and prosody. Its expanded scalar grid combines $10^{-6},10^{-5.5},\ldots,10^3$ with every 1-grid and 5-grid value. For each rule and construct, the equal-penalty winner is additionally refined over nine logarithmically spaced points from $\alpha/\sqrt{10}$ to $\alpha\sqrt{10}$, together with the expanded candidates. The banded search includes that refined equal-penalty solution as a fallback. Relative feature gains $r^{1/4}$ for XEUS and $r^{-1/4}$ for XLS-R yield penalties $\alpha/\sqrt r$ and $\alpha\sqrt r$ in the original standardized coordinates. 3RB-C selects construct Spearman, breaking ties by lower MAE, equal penalties, then larger scalar penalty; 3RB-A selects accuracy MAE, breaking ties by equal penalties, then larger scalar penalty. Accuracy has no banded search and retains its corresponding refined scalar-ridge head. The broadened scalar grid and unequal-penalty search change together, so the comparison does not isolate banding alone. The external fluency leader 3RB-C-aesopocean uses scalar $\alpha=10$ and ratio $r=4$ (XEUS/XLS-R penalties 5 and 20); the prosody-correlation leader 3RB-A-koreanaesop uses $\alpha=5$, $r=4$ (penalties 2.5 and 10). These settings were selected on public development data.

All 35 configurations completed all 305 recordings. The original 1G-S/1G-AMAE/5G-AMAE arrays are unchanged; the 14 3RB variants add the exact evaluated checkpoints. No private ratings entered fitting, tuning, or calibration. The main table reports descriptive external maxima and ties within .001, whereas the paired analysis below separately evaluates uncertainty. Full-precision two- and three-rater statistics are supplied as aggregate CSV files.
\clearpage
\begin{landscape}
\begingroup\fontsize{6.5}{7}\selectfont\setlength{\tabcolsep}{2.5pt}\setlength{\LTcapwidth}{\linewidth}\renewcommand{\arraystretch}{0.95}
\begin{longtable}{@{}l*{10}{c}@{}}
\caption{All families and conditions ($N=305$); marginal 95\% CIs.}\label{tab:all-constructs}\\
\endfirsthead
\multicolumn{11}{l}{\textit{Family comparisons (continued)}}\\\endhead
\bottomrule\endfoot

\multicolumn{11}{l}{\textbf{Accuracy}}\\*
\toprule
Human consensus & \multicolumn{5}{c}{Spearman 0.809 [0.769, 0.842]} & \multicolumn{5}{c}{QWK 0.704 [0.663, 0.743]}\\
Azure & \multicolumn{5}{c}{Spearman 0.761 [0.709, 0.804]} & \multicolumn{5}{c}{QWK 0.161 [0.129, 0.196]}\\
\midrule
Family & \multicolumn{2}{c}{1G-S} & \multicolumn{2}{c}{1G-AMAE} & \multicolumn{2}{c}{5G-AMAE} & \multicolumn{2}{c}{3RB-C} & \multicolumn{2}{c}{3RB-A}\\
 & $\rho$ & QWK & $\rho$ & QWK & $\rho$ & QWK & $\rho$ & QWK & $\rho$ & QWK\\
\midrule
speechocean & \shortstack{0.856\\{}[0.821,0.882]} & \shortstack{0.502\\{}[0.451,0.549]} & \shortstack{0.856\\{}[0.821,0.882]} & \shortstack{0.502\\{}[0.451,0.549]} & \shortstack{0.844\\{}[0.807,0.872]} & \shortstack{0.546\\{}[0.494,0.592]} & \shortstack{0.844\\{}[0.807,0.872]} & \shortstack{0.546\\{}[0.494,0.592]} & \shortstack{0.844\\{}[0.807,0.872]} & \shortstack{0.546\\{}[0.494,0.592]}\\
korean & \shortstack{0.856\\{}[0.822,0.882]} & \shortstack{0.702\\{}[0.662,0.739]} & \shortstack{0.856\\{}[0.822,0.882]} & \shortstack{0.702\\{}[0.662,0.739]} & \shortstack{0.853\\{}[0.820,0.881]} & \shortstack{0.672\\{}[0.631,0.711]} & \shortstack{0.855\\{}[0.822,0.882]} & \shortstack{0.692\\{}[0.652,0.730]} & \shortstack{0.853\\{}[0.819,0.880]} & \shortstack{0.658\\{}[0.617,0.697]}\\
aesop & \shortstack{0.852\\{}[0.818,0.878]} & \shortstack{0.241\\{}[0.215,0.265]} & \shortstack{0.864\\{}[0.832,0.888]} & \shortstack{0.239\\{}[0.214,0.264]} & \shortstack{0.863\\{}[0.830,0.887]} & \shortstack{0.239\\{}[0.214,0.264]} & \shortstack{0.849\\{}[0.814,0.875]} & \shortstack{0.238\\{}[0.212,0.263]} & \shortstack{0.862\\{}[0.830,0.886]} & \shortstack{0.243\\{}[0.217,0.267]}\\
koreanocean & \shortstack{0.853\\{}[0.816,0.882]} & \shortstack{0.821\\{}[0.789,0.847]} & \shortstack{0.841\\{}[0.802,0.871]} & \shortstack{0.798\\{}[0.764,0.827]} & \shortstack{0.849\\{}[0.811,0.879]} & \shortstack{0.819\\{}[0.788,0.846]} & \shortstack{0.852\\{}[0.813,0.881]} & \shortstack{0.821\\{}[0.789,0.847]} & \shortstack{0.841\\{}[0.803,0.872]} & \shortstack{0.802\\{}[0.769,0.831]}\\
aesopocean & \shortstack{0.872\\{}[0.842,0.894]} & \shortstack{0.598\\{}[0.563,0.631]} & \shortstack{0.865\\{}[0.833,0.890]} & \shortstack{0.628\\{}[0.593,0.662]} & \shortstack{0.872\\{}[0.841,0.894]} & \shortstack{0.607\\{}[0.572,0.640]} & \shortstack{0.871\\{}[0.841,0.894]} & \shortstack{0.596\\{}[0.561,0.629]} & \shortstack{0.871\\{}[0.840,0.894]} & \shortstack{0.612\\{}[0.576,0.645]}\\
koreanaesop & \shortstack{0.867\\{}[0.835,0.891]} & \shortstack{0.477\\{}[0.439,0.512]} & \shortstack{0.867\\{}[0.835,0.891]} & \shortstack{0.477\\{}[0.439,0.512]} & \shortstack{0.867\\{}[0.836,0.891]} & \shortstack{0.472\\{}[0.435,0.508]} & \shortstack{0.867\\{}[0.836,0.891]} & \shortstack{0.477\\{}[0.440,0.513]} & \shortstack{0.867\\{}[0.836,0.891]} & \shortstack{0.475\\{}[0.437,0.510]}\\
aesopkoreanocean & \shortstack{0.870\\{}[0.839,0.894]} & \shortstack{0.585\\{}[0.548,0.620]} & \shortstack{0.862\\{}[0.829,0.888]} & \shortstack{0.576\\{}[0.537,0.613]} & \shortstack{0.857\\{}[0.824,0.884]} & \shortstack{0.575\\{}[0.537,0.611]} & \shortstack{0.869\\{}[0.837,0.893]} & \shortstack{0.579\\{}[0.540,0.614]} & \shortstack{0.863\\{}[0.830,0.889]} & \shortstack{0.577\\{}[0.538,0.613]}\\
\midrule
\multicolumn{11}{l}{\textbf{Fluency}}\\*
\toprule
Human consensus & \multicolumn{5}{c}{Spearman 0.807 [0.768, 0.840]} & \multicolumn{5}{c}{QWK 0.739 [0.695, 0.777]}\\
Azure & \multicolumn{5}{c}{Spearman 0.633 [0.556, 0.698]} & \multicolumn{5}{c}{QWK 0.078 [0.056, 0.107]}\\
\midrule
Family & \multicolumn{2}{c}{1G-S} & \multicolumn{2}{c}{1G-AMAE} & \multicolumn{2}{c}{5G-AMAE} & \multicolumn{2}{c}{3RB-C} & \multicolumn{2}{c}{3RB-A}\\
 & $\rho$ & QWK & $\rho$ & QWK & $\rho$ & QWK & $\rho$ & QWK & $\rho$ & QWK\\
\midrule
speechocean & \shortstack{0.811\\{}[0.763,0.848]} & \shortstack{0.716\\{}[0.672,0.752]} & \shortstack{0.847\\{}[0.806,0.878]} & \shortstack{0.629\\{}[0.585,0.666]} & \shortstack{0.842\\{}[0.801,0.874]} & \shortstack{0.670\\{}[0.627,0.706]} & \shortstack{0.832\\{}[0.790,0.866]} & \shortstack{0.704\\{}[0.662,0.739]} & \shortstack{0.836\\{}[0.795,0.870]} & \shortstack{0.693\\{}[0.651,0.728]}\\
korean & \shortstack{0.860\\{}[0.825,0.888]} & \shortstack{0.495\\{}[0.452,0.535]} & \shortstack{0.860\\{}[0.825,0.888]} & \shortstack{0.495\\{}[0.452,0.535]} & \shortstack{0.859\\{}[0.825,0.886]} & \shortstack{0.501\\{}[0.458,0.538]} & \shortstack{0.864\\{}[0.831,0.890]} & \shortstack{0.536\\{}[0.493,0.576]} & \shortstack{0.859\\{}[0.824,0.886]} & \shortstack{0.503\\{}[0.461,0.540]}\\
aesop & \shortstack{0.850\\{}[0.813,0.879]} & \shortstack{0.276\\{}[0.247,0.305]} & \shortstack{0.855\\{}[0.819,0.883]} & \shortstack{0.312\\{}[0.279,0.342]} & \shortstack{0.859\\{}[0.823,0.885]} & \shortstack{0.315\\{}[0.282,0.346]} & \shortstack{0.855\\{}[0.819,0.883]} & \shortstack{0.289\\{}[0.258,0.318]} & \shortstack{0.859\\{}[0.823,0.886]} & \shortstack{0.314\\{}[0.281,0.345]}\\
koreanocean & \shortstack{0.840\\{}[0.799,0.873]} & \shortstack{0.585\\{}[0.537,0.628]} & \shortstack{0.846\\{}[0.808,0.877]} & \shortstack{0.714\\{}[0.672,0.751]} & \shortstack{0.838\\{}[0.799,0.870]} & \shortstack{0.744\\{}[0.702,0.779]} & \shortstack{0.845\\{}[0.805,0.877]} & \shortstack{0.634\\{}[0.588,0.676]} & \shortstack{0.847\\{}[0.809,0.879]} & \shortstack{0.700\\{}[0.657,0.738]}\\
aesopocean & \shortstack{0.874\\{}[0.842,0.899]} & \shortstack{0.613\\{}[0.575,0.647]} & \shortstack{0.874\\{}[0.842,0.899]} & \shortstack{0.613\\{}[0.575,0.647]} & \shortstack{0.862\\{}[0.828,0.890]} & \shortstack{0.612\\{}[0.574,0.647]} & \shortstack{0.883\\{}[0.854,0.906]} & \shortstack{0.751\\{}[0.718,0.779]} & \shortstack{0.881\\{}[0.851,0.905]} & \shortstack{0.742\\{}[0.710,0.771]}\\
koreanaesop & \shortstack{0.876\\{}[0.846,0.900]} & \shortstack{0.462\\{}[0.424,0.496]} & \shortstack{0.876\\{}[0.846,0.900]} & \shortstack{0.462\\{}[0.424,0.496]} & \shortstack{0.878\\{}[0.848,0.901]} & \shortstack{0.469\\{}[0.431,0.502]} & \shortstack{0.874\\{}[0.843,0.898]} & \shortstack{0.452\\{}[0.415,0.486]} & \shortstack{0.878\\{}[0.848,0.902]} & \shortstack{0.488\\{}[0.451,0.522]}\\
aesopkoreanocean & \shortstack{0.874\\{}[0.842,0.898]} & \shortstack{0.527\\{}[0.486,0.564]} & \shortstack{0.877\\{}[0.846,0.902]} & \shortstack{0.577\\{}[0.539,0.612]} & \shortstack{0.871\\{}[0.839,0.897]} & \shortstack{0.591\\{}[0.552,0.626]} & \shortstack{0.880\\{}[0.849,0.903]} & \shortstack{0.629\\{}[0.589,0.664]} & \shortstack{0.877\\{}[0.846,0.902]} & \shortstack{0.577\\{}[0.539,0.612]}\\
\midrule
\multicolumn{11}{l}{\textbf{Prosody}}\\*
\toprule
Human consensus & \multicolumn{5}{c}{Spearman 0.807 [0.768, 0.842]} & \multicolumn{5}{c}{QWK 0.742 [0.701, 0.778]}\\
Azure & \multicolumn{5}{c}{Spearman 0.737 [0.672, 0.792]} & \multicolumn{5}{c}{QWK 0.370 [0.318, 0.421]}\\
\midrule
Family & \multicolumn{2}{c}{1G-S} & \multicolumn{2}{c}{1G-AMAE} & \multicolumn{2}{c}{5G-AMAE} & \multicolumn{2}{c}{3RB-C} & \multicolumn{2}{c}{3RB-A}\\
 & $\rho$ & QWK & $\rho$ & QWK & $\rho$ & QWK & $\rho$ & QWK & $\rho$ & QWK\\
\midrule
speechocean & \shortstack{0.819\\{}[0.776,0.855]} & \shortstack{0.597\\{}[0.551,0.639]} & \shortstack{0.850\\{}[0.813,0.880]} & \shortstack{0.545\\{}[0.501,0.585]} & \shortstack{0.847\\{}[0.809,0.878]} & \shortstack{0.570\\{}[0.525,0.610]} & \shortstack{0.839\\{}[0.800,0.871]} & \shortstack{0.588\\{}[0.543,0.628]} & \shortstack{0.843\\{}[0.804,0.874]} & \shortstack{0.581\\{}[0.535,0.622]}\\
korean & \shortstack{0.874\\{}[0.842,0.899]} & \shortstack{0.489\\{}[0.447,0.526]} & \shortstack{0.874\\{}[0.842,0.899]} & \shortstack{0.489\\{}[0.447,0.526]} & \shortstack{0.872\\{}[0.840,0.896]} & \shortstack{0.493\\{}[0.452,0.531]} & \shortstack{0.878\\{}[0.847,0.902]} & \shortstack{0.530\\{}[0.487,0.568]} & \shortstack{0.871\\{}[0.839,0.895]} & \shortstack{0.495\\{}[0.454,0.533]}\\
aesop & \shortstack{0.850\\{}[0.814,0.878]} & \shortstack{0.261\\{}[0.232,0.288]} & \shortstack{0.855\\{}[0.820,0.882]} & \shortstack{0.280\\{}[0.249,0.309]} & \shortstack{0.858\\{}[0.824,0.884]} & \shortstack{0.283\\{}[0.252,0.312]} & \shortstack{0.846\\{}[0.810,0.875]} & \shortstack{0.255\\{}[0.226,0.282]} & \shortstack{0.863\\{}[0.830,0.889]} & \shortstack{0.270\\{}[0.240,0.298]}\\
koreanocean & \shortstack{0.850\\{}[0.812,0.881]} & \shortstack{0.546\\{}[0.501,0.586]} & \shortstack{0.851\\{}[0.815,0.882]} & \shortstack{0.641\\{}[0.599,0.680]} & \shortstack{0.840\\{}[0.802,0.872]} & \shortstack{0.675\\{}[0.632,0.712]} & \shortstack{0.856\\{}[0.818,0.886]} & \shortstack{0.584\\{}[0.540,0.624]} & \shortstack{0.854\\{}[0.817,0.885]} & \shortstack{0.629\\{}[0.586,0.668]}\\
aesopocean & \shortstack{0.881\\{}[0.850,0.905]} & \shortstack{0.541\\{}[0.502,0.576]} & \shortstack{0.870\\{}[0.838,0.895]} & \shortstack{0.497\\{}[0.457,0.531]} & \shortstack{0.861\\{}[0.827,0.888]} & \shortstack{0.481\\{}[0.440,0.516]} & \shortstack{0.865\\{}[0.831,0.892]} & \shortstack{0.475\\{}[0.434,0.509]} & \shortstack{0.882\\{}[0.852,0.905]} & \shortstack{0.604\\{}[0.566,0.637]}\\
koreanaesop & \shortstack{0.883\\{}[0.852,0.906]} & \shortstack{0.431\\{}[0.394,0.465]} & \shortstack{0.883\\{}[0.852,0.906]} & \shortstack{0.431\\{}[0.394,0.465]} & \shortstack{0.883\\{}[0.853,0.906]} & \shortstack{0.433\\{}[0.397,0.468]} & \shortstack{0.883\\{}[0.852,0.906]} & \shortstack{0.431\\{}[0.394,0.465]} & \shortstack{0.889\\{}[0.860,0.911]} & \shortstack{0.446\\{}[0.409,0.479]}\\
aesopkoreanocean & \shortstack{0.881\\{}[0.850,0.906]} & \shortstack{0.480\\{}[0.440,0.516]} & \shortstack{0.879\\{}[0.847,0.904]} & \shortstack{0.505\\{}[0.467,0.541]} & \shortstack{0.872\\{}[0.839,0.898]} & \shortstack{0.518\\{}[0.479,0.554]} & \shortstack{0.882\\{}[0.850,0.906]} & \shortstack{0.493\\{}[0.454,0.529]} & \shortstack{0.879\\{}[0.847,0.904]} & \shortstack{0.505\\{}[0.467,0.541]}\\
\midrule
\end{longtable}
\endgroup\end{landscape}

\begin{landscape}
\section{Paired differences and score distributions}
The same 305 recordings were jointly resampled 1,200,000 times per seed (42 and 2026), treating each unique recording URL as one speaker. Each replicate recomputed agreement against the same three two-rater reference pairs. Bonferroni intervals use percentile endpoints $0.025/510$ and $1-0.025/510$, covering 35 models versus human consensus and Azure, plus Azure versus human consensus, for three constructs and two metrics, and 84 banded-versus-previous-leader contrasts. Seven of 510 adjusted decisions differ between seeds and are marked with a dagger; these borderline comparisons do not establish a direction. The table reports seed-42 intervals; full-precision results for both seeds are released. Approximately 59 draws fall in each adjusted tail, so endpoints remain Monte Carlo approximations. Correction covers these contrasts, not the preceding model search or all earlier experiments; inference conditions on the observed panel and fitted models. No private scores were used for fitting or calibration.
\begingroup\scriptsize\setlength{\tabcolsep}{3pt}\setlength{\LTcapwidth}{\linewidth}\renewcommand{\arraystretch}{1.0}\setlength{\LTpost}{0pt}
\begin{longtable}{@{}lllrrrrrrrrr@{}}
\caption{Paired differences for all 1G-S/1G-AMAE/5G-AMAE/3RB configurations and constructs.}\label{tab:paired-all}\\
\toprule & & & \multicolumn{3}{c}{Accuracy} & \multicolumn{3}{c}{Fluency} & \multicolumn{3}{c}{Prosody}\\\cmidrule(lr){4-6}\cmidrule(lr){7-9}\cmidrule(lr){10-12}System & Reference & Metric & Diff. & 95\% CI & Adjusted CI & Diff. & 95\% CI & Adjusted CI & Diff. & 95\% CI & Adjusted CI\\\midrule\endfirsthead
\toprule & & & \multicolumn{3}{c}{Accuracy} & \multicolumn{3}{c}{Fluency} & \multicolumn{3}{c}{Prosody}\\\cmidrule(lr){4-6}\cmidrule(lr){7-9}\cmidrule(lr){10-12}System & Reference & Metric & Diff. & 95\% CI & Adjusted CI & Diff. & 95\% CI & Adjusted CI & Diff. & 95\% CI & Adjusted CI\\\midrule\endhead
\bottomrule\endfoot
\multicolumn{12}{l}{\textbf{Azure benchmark (independent of selection rule)}}\\*
azure & Human & $\rho$ & -0.049 & [-0.096, -0.005] & [-0.148, 0.038] & -0.174 & [-0.243, -0.110] & [-0.319, -0.052] & -0.070 & [-0.125, -0.022] & [-0.186, 0.019]\\
azure & Human & QWK & -0.543 & [-0.580, -0.504] & [-0.613, -0.464] & -0.661 & [-0.697, -0.620] & [-0.730, -0.578] & -0.372 & [-0.419, -0.325] & [-0.463, -0.277]\\
\midrule\multicolumn{12}{l}{\textbf{1G-S: 1-grid construct-wise correlation}}\\*
speechocean & Human & $\rho$ & +0.046 & [0.016, 0.076] & [-0.016, 0.109] & +0.004 & [-0.036, 0.041] & [-0.080, 0.076] & +0.012 & [-0.023, 0.044] & [-0.061, 0.076]\\
speechocean & Human & QWK & -0.202 & [-0.249, -0.155] & [-0.294, -0.106] & -0.023 & [-0.067, 0.022] & [-0.112, 0.066] & -0.144 & [-0.188, -0.101] & [-0.231, -0.057]\\
speechocean & Azure & $\rho$ & +0.095 & [0.054, 0.139] & [0.015, 0.189] & +0.178 & [0.113, 0.247] & [0.052, 0.323] & +0.082 & [0.033, 0.135] & [-0.011, 0.197]\\
speechocean & Azure & QWK & +0.341 & [0.299, 0.381] & [0.258, 0.419] & +0.638 & [0.601, 0.671] & [0.561, 0.702] & +0.228 & [0.173, 0.282] & [0.117, 0.334]\\
\addlinespace[2pt]
korean & Human & $\rho$ & +0.046 & [0.018, 0.075] & [-0.012, 0.108] & +0.054 & [0.022, 0.085] & [-0.011, 0.120] & +0.067 & [0.039, 0.096] & [0.011, 0.128]\\
korean & Human & QWK & -0.002 & [-0.041, 0.037] & [-0.080, 0.076] & -0.244 & [-0.281, -0.208] & [-0.316, -0.171] & -0.253 & [-0.287, -0.219] & [-0.319, -0.187]\\
korean & Azure & $\rho$ & +0.095 & [0.055, 0.139] & [0.017, 0.189] & +0.228 & [0.164, 0.297] & [0.107, 0.373] & +0.137 & [0.098, 0.183] & [0.066, 0.237]\\
korean & Azure & QWK & +0.541 & [0.500, 0.580] & [0.460, 0.617] & +0.417 & [0.379, 0.452] & [0.341, 0.486] & +0.119 & [0.077, 0.160] & [0.034, 0.199]\\
\addlinespace[2pt]
aesop & Human & $\rho$ & +0.043 & [0.011, 0.074] & [-0.021, 0.110] & +0.044 & [0.009, 0.077] & [-0.026, 0.112] & +0.043 & [0.012, 0.073] & [-0.020, 0.106]\\
aesop & Human & QWK & -0.463 & [-0.493, -0.431] & [-0.521, -0.396] & -0.463 & [-0.494, -0.430] & [-0.524, -0.396] & -0.481 & [-0.510, -0.451] & [-0.536, -0.420]\\
aesop & Azure & $\rho$ & +0.092 & [0.054, 0.132] & [0.020, 0.179] & +0.218 & [0.157, 0.284] & [0.104, 0.358] & +0.113 & [0.073, 0.159] & [0.040, 0.213]\\
aesop & Azure & QWK & +0.080 & [0.053, 0.105] & [0.023, 0.128] & +0.198 & [0.174, 0.221] & [0.150, 0.242] & -0.109 & [-0.147, -0.072] & [-0.186, -0.037]\\
\addlinespace[2pt]
koreanocean & Human & $\rho$ & +0.044 & [0.013, 0.074] & [-0.021, 0.108] & +0.033 & [-0.004, 0.068] & [-0.045, 0.104] & +0.042 & [0.010, 0.074] & [-0.026, 0.106]\\
koreanocean & Human & QWK & +0.116 & [0.082, 0.152] & [0.048, 0.191] & -0.154 & [-0.195, -0.114] & [-0.236, -0.076] & -0.196 & [-0.232, -0.161] & [-0.268, -0.126]\\
koreanocean & Azure & $\rho$ & +0.093 & [0.051, 0.137] & [0.012, 0.187] & +0.207 & [0.140, 0.278] & [0.079, 0.359] & +0.113 & [0.069, 0.163] & [0.032, 0.222]\\
koreanocean & Azure & QWK & +0.660 & [0.626, 0.690] & [0.591, 0.718] & +0.507 & [0.464, 0.546] & [0.419, 0.584] & +0.176 & [0.128, 0.223] & [0.077, 0.269]\\
\addlinespace[2pt]
aesopocean & Human & $\rho$ & +0.062 & [0.033, 0.093] & [0.003, 0.127] & +0.068 & [0.038, 0.098] & [0.008, 0.131] & +0.073 & [0.048, 0.100] & [0.024, 0.129]\\
aesopocean & Human & QWK & -0.106 & [-0.140, -0.071] & [-0.172, -0.035] & -0.126 & [-0.159, -0.093] & [-0.192, -0.059] & -0.201 & [-0.232, -0.170] & [-0.262, -0.138]\\
aesopocean & Azure & $\rho$ & +0.111 & [0.071, 0.155] & [0.034, 0.206] & +0.242 & [0.180, 0.309] & [0.126, 0.385] & +0.144 & [0.099, 0.195] & [0.063, 0.256]\\
aesopocean & Azure & QWK & +0.437 & [0.402, 0.470] & [0.368, 0.503] & +0.535 & [0.501, 0.566] & [0.465, 0.595] & +0.171 & [0.128, 0.214] & [0.084, 0.254]\\
\addlinespace[2pt]
koreanaesop & Human & $\rho$ & +0.057 & [0.028, 0.088] & [-0.003, 0.121] & +0.069 & [0.040, 0.099] & [0.013, 0.133] & +0.076 & [0.049, 0.104] & [0.023, 0.135]\\
koreanaesop & Human & QWK & -0.227 & [-0.261, -0.193] & [-0.293, -0.157] & -0.277 & [-0.311, -0.243] & [-0.344, -0.210] & -0.311 & [-0.342, -0.279] & [-0.372, -0.247]\\
koreanaesop & Azure & $\rho$ & +0.106 & [0.066, 0.150] & [0.028, 0.201] & +0.243 & [0.182, 0.311] & [0.128, 0.385] & +0.146 & [0.105, 0.194] & [0.073, 0.250]\\
koreanaesop & Azure & QWK & +0.316 & [0.281, 0.350] & [0.247, 0.383] & +0.384 & [0.350, 0.415] & [0.317, 0.447] & +0.062 & [0.022, 0.100] & [-0.018, 0.136]\\
\addlinespace[2pt]
aesopkoreanocean & Human & $\rho$ & +0.061 & [0.032, 0.091] & [0.002, 0.125] & +0.067 & [0.037, 0.098] & [0.007, 0.131] & +0.074 & [0.047, 0.101] & [0.020, 0.132]\\
aesopkoreanocean & Human & QWK & -0.119 & [-0.153, -0.084] & [-0.186, -0.048] & -0.212 & [-0.247, -0.178] & [-0.281, -0.145] & -0.262 & [-0.293, -0.231] & [-0.323, -0.200]\\
aesopkoreanocean & Azure & $\rho$ & +0.110 & [0.070, 0.154] & [0.033, 0.204] & +0.241 & [0.178, 0.311] & [0.121, 0.388] & +0.144 & [0.101, 0.194] & [0.067, 0.253]\\
aesopkoreanocean & Azure & QWK & +0.425 & [0.388, 0.459] & [0.353, 0.492] & +0.449 & [0.412, 0.482] & [0.375, 0.514] & +0.111 & [0.068, 0.151] & [0.025, 0.191]\\
\addlinespace[2pt]
\midrule\multicolumn{12}{l}{\textbf{1G-AMAE: 1-grid pronunciation-accuracy error}}\\*
speechocean & Human & $\rho$ & +0.046 & [0.016, 0.076] & [-0.016, 0.109] & +0.040 & [0.006, 0.073] & [-0.032, 0.106] & +0.043 & [0.012, 0.072] & [-0.021, 0.103]\\
speechocean & Human & QWK & -0.202 & [-0.249, -0.155] & [-0.294, -0.106] & -0.111 & [-0.156, -0.064] & [-0.200, -0.016] & -0.196 & [-0.240, -0.151] & [-0.281, -0.104]\\
speechocean & Azure & $\rho$ & +0.095 & [0.054, 0.139] & [0.015, 0.189] & +0.214 & [0.150, 0.283] & [0.092, 0.359] & +0.113 & [0.066, 0.166] & [0.025, 0.229]\\
speechocean & Azure & QWK & +0.341 & [0.299, 0.381] & [0.258, 0.419] & +0.550 & [0.511, 0.586] & [0.470, 0.620] & +0.176 & [0.119, 0.232] & [0.062, 0.287]\\
\addlinespace[2pt]
korean & Human & $\rho$ & +0.046 & [0.018, 0.075] & [-0.012, 0.108] & +0.054 & [0.022, 0.085] & [-0.011, 0.120] & +0.067 & [0.039, 0.096] & [0.011, 0.128]\\
korean & Human & QWK & -0.002 & [-0.041, 0.037] & [-0.080, 0.076] & -0.244 & [-0.281, -0.208] & [-0.316, -0.171] & -0.253 & [-0.287, -0.219] & [-0.319, -0.187]\\
korean & Azure & $\rho$ & +0.095 & [0.055, 0.139] & [0.017, 0.189] & +0.228 & [0.164, 0.297] & [0.107, 0.373] & +0.137 & [0.098, 0.183] & [0.066, 0.237]\\
korean & Azure & QWK & +0.541 & [0.500, 0.580] & [0.460, 0.617] & +0.417 & [0.379, 0.452] & [0.341, 0.486] & +0.119 & [0.077, 0.160] & [0.034, 0.199]\\
\addlinespace[2pt]
aesop & Human & $\rho$ & +0.054 & [0.025, 0.085] & [-0.006, 0.119] & +0.048 & [0.015, 0.081] & [-0.020, 0.117] & +0.048 & [0.018, 0.078] & [-0.012, 0.110]\\
aesop & Human & QWK & -0.465 & [-0.493, -0.434] & [-0.520, -0.399] & -0.428 & [-0.459, -0.395] & [-0.489, -0.362] & -0.462 & [-0.491, -0.432] & [-0.518, -0.401]\\
aesop & Azure & $\rho$ & +0.103 & [0.065, 0.146] & [0.029, 0.195] & +0.222 & [0.163, 0.288] & [0.109, 0.360] & +0.119 & [0.077, 0.166] & [0.042, 0.220]\\
aesop & Azure & QWK & +0.079 & [0.052, 0.103] & [0.022, 0.126] & +0.233 & [0.206, 0.259] & [0.179, 0.285] & -0.090 & [-0.127, -0.053] & [-0.166, -0.018]\\
\addlinespace[2pt]
koreanocean & Human & $\rho$ & +0.031 & [-0.000, 0.062] & [-0.035, 0.094] & +0.039 & [0.005, 0.072] & [-0.031, 0.106] & +0.044 & [0.014, 0.073] & [-0.019, 0.104]\\
koreanocean & Human & QWK & +0.093 & [0.059, 0.129] & [0.024, 0.166] & -0.025 & [-0.063, 0.012] & [-0.101, 0.049] & -0.101 & [-0.135, -0.067] & [-0.169, -0.034]\\
koreanocean & Azure & $\rho$ & +0.080 & [0.037, 0.125] & [-0.003, 0.176] & +0.213 & [0.148, 0.283] & [0.090, 0.362] & +0.114 & [0.071, 0.164] & [0.035, 0.222]\\
koreanocean & Azure & QWK & +0.637 & [0.600, 0.671] & [0.563, 0.702] & +0.636 & [0.595, 0.672] & [0.554, 0.705] & +0.272 & [0.225, 0.317] & [0.177, 0.361]\\
\addlinespace[2pt]
aesopocean & Human & $\rho$ & +0.056 & [0.026, 0.086] & [-0.005, 0.120] & +0.068 & [0.038, 0.098] & [0.008, 0.131] & +0.063 & [0.037, 0.089] & [0.010, 0.119]\\
aesopocean & Human & QWK & -0.076 & [-0.109, -0.041] & [-0.142, -0.004] & -0.126 & [-0.159, -0.093] & [-0.192, -0.059] & -0.245 & [-0.276, -0.214] & [-0.305, -0.183]\\
aesopocean & Azure & $\rho$ & +0.105 & [0.063, 0.150] & [0.024, 0.203] & +0.242 & [0.180, 0.309] & [0.126, 0.385] & +0.133 & [0.089, 0.184] & [0.052, 0.245]\\
aesopocean & Azure & QWK & +0.468 & [0.432, 0.502] & [0.396, 0.534] & +0.535 & [0.501, 0.566] & [0.465, 0.595] & +0.127 & [0.085, 0.168] & [0.042, 0.208]\\
\addlinespace[2pt]
koreanaesop & Human & $\rho$ & +0.057 & [0.028, 0.088] & [-0.003, 0.121] & +0.069 & [0.040, 0.099] & [0.013, 0.133] & +0.076 & [0.049, 0.104] & [0.023, 0.135]\\
koreanaesop & Human & QWK & -0.227 & [-0.261, -0.193] & [-0.293, -0.157] & -0.277 & [-0.311, -0.243] & [-0.344, -0.210] & -0.311 & [-0.342, -0.279] & [-0.372, -0.247]\\
koreanaesop & Azure & $\rho$ & +0.106 & [0.066, 0.150] & [0.028, 0.201] & +0.243 & [0.182, 0.311] & [0.128, 0.385] & +0.146 & [0.105, 0.194] & [0.073, 0.250]\\
koreanaesop & Azure & QWK & +0.316 & [0.281, 0.350] & [0.247, 0.383] & +0.384 & [0.350, 0.415] & [0.317, 0.447] & +0.062 & [0.022, 0.100] & [-0.018, 0.136]\\
\addlinespace[2pt]
aesopkoreanocean & Human & $\rho$ & +0.052 & [0.022, 0.083] & [-0.009, 0.117] & +0.070 & [0.040, 0.101] & [0.010, 0.135] & +0.072 & [0.045, 0.099] & [0.017, 0.130]\\
aesopkoreanocean & Human & QWK & -0.128 & [-0.162, -0.094] & [-0.195, -0.058] & -0.162 & [-0.196, -0.128] & [-0.230, -0.094] & -0.237 & [-0.268, -0.205] & [-0.298, -0.175]\\
aesopkoreanocean & Azure & $\rho$ & +0.101 & [0.060, 0.146] & [0.023, 0.198] & +0.244 & [0.182, 0.313] & [0.127, 0.390] & +0.142 & [0.099, 0.191] & [0.066, 0.249]\\
aesopkoreanocean & Azure & QWK & +0.415 & [0.378, 0.451] & [0.342, 0.485] & +0.499 & [0.463, 0.532] & [0.426, 0.563] & +0.136 & [0.095, 0.175] & [0.053, 0.213]\\
\addlinespace[2pt]
\midrule\multicolumn{12}{l}{\textbf{5G-AMAE: 5-grid pronunciation-accuracy error}}\\*
speechocean & Human & $\rho$ & +0.035 & [0.004, 0.065] & [-0.029, 0.097] & +0.035 & [-0.000, 0.069] & [-0.039, 0.102] & +0.040 & [0.009, 0.069] & [-0.024, 0.099]\\
speechocean & Human & QWK & -0.158 & [-0.206, -0.110] & [-0.253, -0.061] & -0.069 & [-0.113, -0.024] & [-0.156, 0.023] & -0.172 & [-0.215, -0.128] & [-0.255, -0.083]\\
speechocean & Azure & $\rho$ & +0.083 & [0.041, 0.128] & [-0.000, 0.178]$^{\dagger}$ & +0.209 & [0.145, 0.278] & [0.086, 0.354] & +0.110 & [0.063, 0.163] & [0.021, 0.225]\\
speechocean & Azure & QWK & +0.385 & [0.341, 0.426] & [0.298, 0.464] & +0.592 & [0.553, 0.627] & [0.513, 0.660] & +0.200 & [0.145, 0.255] & [0.088, 0.308]\\
\addlinespace[2pt]
korean & Human & $\rho$ & +0.044 & [0.015, 0.073] & [-0.015, 0.106] & +0.053 & [0.021, 0.084] & [-0.012, 0.119] & +0.064 & [0.036, 0.093] & [0.007, 0.127]\\
korean & Human & QWK & -0.032 & [-0.071, 0.008] & [-0.111, 0.047] & -0.239 & [-0.276, -0.202] & [-0.311, -0.166] & -0.249 & [-0.283, -0.214] & [-0.316, -0.181]\\
korean & Azure & $\rho$ & +0.093 & [0.052, 0.137] & [0.013, 0.187] & +0.227 & [0.163, 0.296] & [0.106, 0.372] & +0.135 & [0.094, 0.181] & [0.061, 0.236]\\
korean & Azure & QWK & +0.511 & [0.470, 0.551] & [0.428, 0.589] & +0.422 & [0.385, 0.457] & [0.347, 0.491] & +0.124 & [0.082, 0.164] & [0.039, 0.203]\\
\addlinespace[2pt]
aesop & Human & $\rho$ & +0.053 & [0.024, 0.083] & [-0.007, 0.118] & +0.052 & [0.019, 0.085] & [-0.015, 0.120] & +0.051 & [0.022, 0.080] & [-0.008, 0.112]\\
aesop & Human & QWK & -0.465 & [-0.493, -0.434] & [-0.520, -0.399] & -0.424 & [-0.455, -0.391] & [-0.485, -0.358] & -0.459 & [-0.488, -0.429] & [-0.515, -0.397]\\
aesop & Azure & $\rho$ & +0.102 & [0.063, 0.145] & [0.027, 0.193] & +0.226 & [0.166, 0.291] & [0.113, 0.364] & +0.122 & [0.080, 0.168] & [0.046, 0.223]\\
aesop & Azure & QWK & +0.079 & [0.052, 0.103] & [0.022, 0.126] & +0.237 & [0.209, 0.263] & [0.183, 0.288] & -0.087 & [-0.124, -0.050] & [-0.162, -0.016]\\
\addlinespace[2pt]
koreanocean & Human & $\rho$ & +0.040 & [0.009, 0.071] & [-0.026, 0.103] & +0.032 & [-0.002, 0.064] & [-0.039, 0.099] & +0.033 & [0.002, 0.062] & [-0.031, 0.093]\\
koreanocean & Human & QWK & +0.115 & [0.080, 0.151] & [0.045, 0.190] & +0.005 & [-0.033, 0.042] & [-0.070, 0.080] & -0.067 & [-0.102, -0.033] & [-0.137, 0.001]\\
koreanocean & Azure & $\rho$ & +0.089 & [0.047, 0.134] & [0.007, 0.184] & +0.206 & [0.141, 0.276] & [0.082, 0.354] & +0.103 & [0.059, 0.153] & [0.023, 0.210]\\
koreanocean & Azure & QWK & +0.659 & [0.624, 0.690] & [0.587, 0.719] & +0.666 & [0.627, 0.701] & [0.586, 0.733] & +0.305 & [0.258, 0.351] & [0.210, 0.395]\\
\addlinespace[2pt]
aesopocean & Human & $\rho$ & +0.062 & [0.033, 0.092] & [0.003, 0.126] & +0.056 & [0.024, 0.087] & [-0.008, 0.121] & +0.054 & [0.026, 0.081] & [-0.003, 0.111]\\
aesopocean & Human & QWK & -0.097 & [-0.131, -0.063] & [-0.163, -0.027] & -0.127 & [-0.161, -0.093] & [-0.194, -0.059] & -0.261 & [-0.292, -0.230] & [-0.322, -0.199]\\
aesopocean & Azure & $\rho$ & +0.111 & [0.070, 0.156] & [0.033, 0.208] & +0.230 & [0.168, 0.297] & [0.114, 0.372] & +0.124 & [0.079, 0.175] & [0.041, 0.236]\\
aesopocean & Azure & QWK & +0.446 & [0.411, 0.480] & [0.376, 0.512] & +0.534 & [0.500, 0.565] & [0.464, 0.595] & +0.111 & [0.068, 0.152] & [0.024, 0.192]\\
\addlinespace[2pt]
koreanaesop & Human & $\rho$ & +0.058 & [0.029, 0.088] & [-0.001, 0.121]$^{\dagger}$ & +0.071 & [0.042, 0.101] & [0.014, 0.135] & +0.076 & [0.049, 0.104] & [0.023, 0.136]\\
koreanaesop & Human & QWK & -0.232 & [-0.265, -0.198] & [-0.296, -0.163] & -0.271 & [-0.304, -0.237] & [-0.337, -0.202] & -0.308 & [-0.340, -0.277] & [-0.370, -0.245]\\
koreanaesop & Azure & $\rho$ & +0.107 & [0.067, 0.151] & [0.029, 0.202] & +0.245 & [0.183, 0.312] & [0.129, 0.388] & +0.147 & [0.105, 0.194] & [0.072, 0.250]\\
koreanaesop & Azure & QWK & +0.312 & [0.277, 0.345] & [0.243, 0.378] & +0.390 & [0.356, 0.422] & [0.323, 0.453] & +0.064 & [0.025, 0.102] & [-0.015, 0.138]\\
\addlinespace[2pt]
aesopkoreanocean & Human & $\rho$ & +0.048 & [0.018, 0.078] & [-0.014, 0.112] & +0.065 & [0.034, 0.096] & [0.004, 0.130] & +0.065 & [0.038, 0.092] & [0.009, 0.123]\\
aesopkoreanocean & Human & QWK & -0.129 & [-0.163, -0.094] & [-0.196, -0.059] & -0.148 & [-0.183, -0.113] & [-0.217, -0.078] & -0.224 & [-0.256, -0.192] & [-0.288, -0.160]\\
aesopkoreanocean & Azure & $\rho$ & +0.097 & [0.056, 0.141] & [0.018, 0.193] & +0.239 & [0.176, 0.307] & [0.121, 0.384] & +0.135 & [0.093, 0.184] & [0.058, 0.242]\\
aesopkoreanocean & Azure & QWK & +0.415 & [0.377, 0.450] & [0.341, 0.485] & +0.513 & [0.477, 0.546] & [0.440, 0.578] & +0.148 & [0.107, 0.188] & [0.064, 0.227]\\
\addlinespace[2pt]
\midrule\multicolumn{12}{l}{\textbf{3RB-C: three-ratio banded, construct-wise correlation}}\\*
speechocean & Human & $\rho$ & +0.035 & [0.004, 0.065] & [-0.029, 0.097] & +0.025 & [-0.011, 0.060] & [-0.052, 0.093] & +0.032 & [0.000, 0.062] & [-0.035, 0.092]\\
speechocean & Human & QWK & -0.158 & [-0.206, -0.110] & [-0.253, -0.061] & -0.035 & [-0.079, 0.009] & [-0.123, 0.053] & -0.154 & [-0.196, -0.111] & [-0.237, -0.067]\\
speechocean & Azure & $\rho$ & +0.083 & [0.041, 0.128] & [-0.000, 0.178]$^{\dagger}$ & +0.199 & [0.135, 0.268] & [0.076, 0.345] & +0.102 & [0.054, 0.155] & [0.012, 0.217]\\
speechocean & Azure & QWK & +0.385 & [0.341, 0.426] & [0.298, 0.464] & +0.626 & [0.589, 0.659] & [0.549, 0.691] & +0.218 & [0.163, 0.273] & [0.108, 0.325]\\
\addlinespace[2pt]
korean & Human & $\rho$ & +0.046 & [0.017, 0.075] & [-0.012, 0.108] & +0.058 & [0.027, 0.088] & [-0.003, 0.122] & +0.071 & [0.044, 0.098] & [0.017, 0.130]\\
korean & Human & QWK & -0.012 & [-0.051, 0.027] & [-0.090, 0.067] & -0.203 & [-0.241, -0.166] & [-0.278, -0.130] & -0.211 & [-0.245, -0.178] & [-0.279, -0.145]\\
korean & Azure & $\rho$ & +0.095 & [0.054, 0.139] & [0.016, 0.189] & +0.232 & [0.169, 0.300] & [0.112, 0.377] & +0.141 & [0.100, 0.189] & [0.068, 0.246]\\
korean & Azure & QWK & +0.531 & [0.490, 0.571] & [0.450, 0.608] & +0.458 & [0.419, 0.493] & [0.380, 0.527] & +0.161 & [0.117, 0.203] & [0.072, 0.245]\\
\addlinespace[2pt]
aesop & Human & $\rho$ & +0.039 & [0.008, 0.071] & [-0.024, 0.106] & +0.049 & [0.015, 0.081] & [-0.019, 0.116] & +0.039 & [0.008, 0.070] & [-0.025, 0.103]\\
aesop & Human & QWK & -0.466 & [-0.496, -0.434] & [-0.523, -0.398] & -0.451 & [-0.482, -0.418] & [-0.512, -0.384] & -0.487 & [-0.515, -0.456] & [-0.542, -0.424]\\
aesop & Azure & $\rho$ & +0.088 & [0.051, 0.128] & [0.017, 0.174] & +0.223 & [0.162, 0.289] & [0.109, 0.363] & +0.110 & [0.069, 0.155] & [0.037, 0.209]\\
aesop & Azure & QWK & +0.077 & [0.051, 0.102] & [0.021, 0.125] & +0.210 & [0.185, 0.234] & [0.160, 0.257] & -0.115 & [-0.153, -0.077] & [-0.192, -0.042]\\
\addlinespace[2pt]
koreanocean & Human & $\rho$ & +0.042 & [0.011, 0.073] & [-0.023, 0.106] & +0.038 & [0.002, 0.073] & [-0.038, 0.109] & +0.048 & [0.017, 0.078] & [-0.017, 0.110]\\
koreanocean & Human & QWK & +0.116 & [0.082, 0.152] & [0.047, 0.191] & -0.105 & [-0.144, -0.066] & [-0.185, -0.028] & -0.158 & [-0.192, -0.124] & [-0.226, -0.091]\\
koreanocean & Azure & $\rho$ & +0.091 & [0.049, 0.136] & [0.010, 0.187] & +0.212 & [0.146, 0.284] & [0.086, 0.364] & +0.119 & [0.075, 0.168] & [0.039, 0.227]\\
koreanocean & Azure & QWK & +0.660 & [0.626, 0.691] & [0.589, 0.719] & +0.556 & [0.513, 0.595] & [0.468, 0.631] & +0.215 & [0.168, 0.260] & [0.119, 0.304]\\
\addlinespace[2pt]
aesopocean & Human & $\rho$ & +0.062 & [0.032, 0.092] & [0.002, 0.127] & +0.077 & [0.048, 0.107] & [0.020, 0.140] & +0.058 & [0.031, 0.085] & [0.004, 0.114]\\
aesopocean & Human & QWK & -0.109 & [-0.142, -0.074] & [-0.175, -0.038] & +0.012 & [-0.022, 0.046] & [-0.055, 0.082] & -0.267 & [-0.298, -0.237] & [-0.327, -0.205]\\
aesopocean & Azure & $\rho$ & +0.111 & [0.071, 0.155] & [0.034, 0.206] & +0.251 & [0.188, 0.319] & [0.134, 0.396] & +0.128 & [0.084, 0.178] & [0.047, 0.239]\\
aesopocean & Azure & QWK & +0.435 & [0.400, 0.468] & [0.366, 0.500] & +0.673 & [0.640, 0.702] & [0.605, 0.729] & +0.105 & [0.063, 0.146] & [0.019, 0.185]\\
\addlinespace[2pt]
koreanaesop & Human & $\rho$ & +0.058 & [0.028, 0.088] & [-0.002, 0.122] & +0.068 & [0.038, 0.097] & [0.010, 0.131] & +0.076 & [0.049, 0.104] & [0.023, 0.135]\\
koreanaesop & Human & QWK & -0.227 & [-0.261, -0.192] & [-0.292, -0.157] & -0.287 & [-0.321, -0.253] & [-0.353, -0.219] & -0.311 & [-0.342, -0.279] & [-0.372, -0.247]\\
koreanaesop & Azure & $\rho$ & +0.107 & [0.066, 0.150] & [0.029, 0.201] & +0.242 & [0.180, 0.309] & [0.126, 0.384] & +0.146 & [0.105, 0.194] & [0.073, 0.250]\\
koreanaesop & Azure & QWK & +0.316 & [0.281, 0.350] & [0.247, 0.383] & +0.374 & [0.340, 0.405] & [0.307, 0.436] & +0.062 & [0.022, 0.100] & [-0.018, 0.136]\\
\addlinespace[2pt]
aesopkoreanocean & Human & $\rho$ & +0.059 & [0.030, 0.090] & [-0.001, 0.123]$^{\dagger}$ & +0.073 & [0.043, 0.104] & [0.013, 0.138] & +0.075 & [0.048, 0.102] & [0.021, 0.133]\\
aesopkoreanocean & Human & QWK & -0.125 & [-0.160, -0.091] & [-0.193, -0.055] & -0.111 & [-0.145, -0.076] & [-0.180, -0.042] & -0.249 & [-0.279, -0.218] & [-0.310, -0.187]\\
aesopkoreanocean & Azure & $\rho$ & +0.108 & [0.068, 0.152] & [0.031, 0.204] & +0.247 & [0.184, 0.316] & [0.129, 0.394] & +0.145 & [0.103, 0.195] & [0.069, 0.253]\\
aesopkoreanocean & Azure & QWK & +0.418 & [0.381, 0.453] & [0.345, 0.486] & +0.551 & [0.514, 0.584] & [0.476, 0.616] & +0.124 & [0.083, 0.164] & [0.040, 0.202]\\
\addlinespace[2pt]
\midrule\multicolumn{12}{l}{\textbf{3RB-A: three-ratio banded, pronunciation-accuracy error}}\\*
speechocean & Human & $\rho$ & +0.035 & [0.004, 0.065] & [-0.029, 0.097] & +0.030 & [-0.006, 0.064] & [-0.045, 0.097] & +0.035 & [0.004, 0.065] & [-0.030, 0.095]\\
speechocean & Human & QWK & -0.158 & [-0.206, -0.110] & [-0.253, -0.061] & -0.046 & [-0.090, -0.001] & [-0.133, 0.044] & -0.161 & [-0.203, -0.117] & [-0.244, -0.073]\\
speechocean & Azure & $\rho$ & +0.083 & [0.041, 0.128] & [-0.000, 0.178]$^{\dagger}$ & +0.204 & [0.139, 0.273] & [0.080, 0.349] & +0.106 & [0.058, 0.159] & [0.016, 0.221]\\
speechocean & Azure & QWK & +0.385 & [0.341, 0.426] & [0.298, 0.464] & +0.615 & [0.577, 0.649] & [0.537, 0.681] & +0.212 & [0.156, 0.266] & [0.101, 0.319]\\
\addlinespace[2pt]
korean & Human & $\rho$ & +0.043 & [0.014, 0.072] & [-0.016, 0.106] & +0.052 & [0.020, 0.084] & [-0.012, 0.118] & +0.063 & [0.035, 0.093] & [0.006, 0.126]\\
korean & Human & QWK & -0.046 & [-0.086, -0.007] & [-0.125, 0.032] & -0.236 & [-0.273, -0.200] & [-0.309, -0.163] & -0.246 & [-0.281, -0.212] & [-0.314, -0.178]\\
korean & Azure & $\rho$ & +0.092 & [0.051, 0.136] & [0.012, 0.186] & +0.226 & [0.162, 0.295] & [0.106, 0.371] & +0.134 & [0.093, 0.181] & [0.060, 0.235]\\
korean & Azure & QWK & +0.497 & [0.455, 0.537] & [0.415, 0.575] & +0.425 & [0.387, 0.460] & [0.350, 0.494] & +0.126 & [0.084, 0.166] & [0.041, 0.205]\\
\addlinespace[2pt]
aesop & Human & $\rho$ & +0.052 & [0.022, 0.083] & [-0.008, 0.118] & +0.052 & [0.019, 0.085] & [-0.015, 0.119] & +0.056 & [0.027, 0.085] & [-0.002, 0.116]\\
aesop & Human & QWK & -0.462 & [-0.490, -0.430] & [-0.518, -0.396] & -0.425 & [-0.457, -0.392] & [-0.487, -0.359] & -0.472 & [-0.500, -0.442] & [-0.526, -0.410]\\
aesop & Azure & $\rho$ & +0.101 & [0.063, 0.143] & [0.028, 0.191] & +0.226 & [0.165, 0.292] & [0.112, 0.367] & +0.126 & [0.084, 0.174] & [0.049, 0.230]\\
aesop & Azure & QWK & +0.082 & [0.055, 0.107] & [0.025, 0.130] & +0.236 & [0.208, 0.263] & [0.181, 0.289] & -0.099 & [-0.137, -0.062] & [-0.176, -0.027]\\
\addlinespace[2pt]
koreanocean & Human & $\rho$ & +0.032 & [-0.000, 0.062] & [-0.035, 0.095] & +0.041 & [0.006, 0.074] & [-0.030, 0.109] & +0.047 & [0.017, 0.076] & [-0.016, 0.108]\\
koreanocean & Human & QWK & +0.098 & [0.063, 0.134] & [0.028, 0.170] & -0.039 & [-0.077, -0.002] & [-0.116, 0.035] & -0.113 & [-0.147, -0.080] & [-0.181, -0.047]\\
koreanocean & Azure & $\rho$ & +0.081 & [0.038, 0.126] & [-0.003, 0.177] & +0.215 & [0.150, 0.285] & [0.092, 0.364] & +0.117 & [0.074, 0.167] & [0.039, 0.224]\\
koreanocean & Azure & QWK & +0.641 & [0.605, 0.675] & [0.568, 0.706] & +0.622 & [0.581, 0.658] & [0.538, 0.692] & +0.259 & [0.212, 0.304] & [0.164, 0.349]\\
\addlinespace[2pt]
aesopocean & Human & $\rho$ & +0.061 & [0.032, 0.092] & [0.002, 0.126] & +0.075 & [0.045, 0.105] & [0.016, 0.139] & +0.074 & [0.049, 0.101] & [0.024, 0.131]\\
aesopocean & Human & QWK & -0.092 & [-0.126, -0.058] & [-0.158, -0.022] & +0.003 & [-0.030, 0.037] & [-0.062, 0.072] & -0.138 & [-0.169, -0.107] & [-0.199, -0.076]\\
aesopocean & Azure & $\rho$ & +0.110 & [0.069, 0.155] & [0.031, 0.207] & +0.249 & [0.187, 0.317] & [0.132, 0.393] & +0.145 & [0.100, 0.196] & [0.064, 0.256]\\
aesopocean & Azure & QWK & +0.451 & [0.416, 0.484] & [0.381, 0.517] & +0.664 & [0.632, 0.693] & [0.598, 0.721] & +0.234 & [0.191, 0.276] & [0.147, 0.318]\\
\addlinespace[2pt]
koreanaesop & Human & $\rho$ & +0.058 & [0.029, 0.088] & [-0.001, 0.121]$^{\dagger}$ & +0.072 & [0.043, 0.102] & [0.015, 0.136] & +0.082 & [0.056, 0.109] & [0.031, 0.141]\\
koreanaesop & Human & QWK & -0.229 & [-0.263, -0.195] & [-0.294, -0.160] & -0.251 & [-0.285, -0.216] & [-0.319, -0.181] & -0.296 & [-0.327, -0.264] & [-0.357, -0.232]\\
koreanaesop & Azure & $\rho$ & +0.107 & [0.067, 0.151] & [0.029, 0.202] & +0.246 & [0.184, 0.313] & [0.130, 0.388] & +0.152 & [0.109, 0.202] & [0.075, 0.261]\\
koreanaesop & Azure & QWK & +0.314 & [0.279, 0.348] & [0.245, 0.380] & +0.410 & [0.376, 0.443] & [0.342, 0.475] & +0.077 & [0.037, 0.115] & [-0.003, 0.152]\\
\addlinespace[2pt]
aesopkoreanocean & Human & $\rho$ & +0.054 & [0.024, 0.084] & [-0.007, 0.118] & +0.070 & [0.040, 0.101] & [0.010, 0.135] & +0.072 & [0.045, 0.099] & [0.017, 0.130]\\
aesopkoreanocean & Human & QWK & -0.127 & [-0.161, -0.093] & [-0.194, -0.057] & -0.162 & [-0.196, -0.128] & [-0.230, -0.094] & -0.237 & [-0.268, -0.205] & [-0.298, -0.175]\\
aesopkoreanocean & Azure & $\rho$ & +0.103 & [0.062, 0.147] & [0.024, 0.199] & +0.244 & [0.182, 0.313] & [0.127, 0.390] & +0.142 & [0.099, 0.191] & [0.066, 0.249]\\
aesopkoreanocean & Azure & QWK & +0.416 & [0.379, 0.452] & [0.343, 0.486] & +0.499 & [0.463, 0.532] & [0.426, 0.563] & +0.136 & [0.095, 0.175] & [0.053, 0.213]\\
\addlinespace[2pt]
\midrule\multicolumn{12}{l}{\textbf{Banded minus previous 1G-S/1G-AMAE/5G-AMAE metric leader}}\\*
\multicolumn{12}{l}{\textbf{3RB-C}}\\*
speechocean & Previous best & $\rho$ & -0.028 & [-0.047, -0.010] & [-0.070, 0.007] & -0.046 & [-0.069, -0.025] & [-0.096, -0.006] & -0.044 & [-0.068, -0.023] & [-0.097, -0.005]\\
speechocean & Previous best & QWK & -0.275 & [-0.314, -0.236] & [-0.350, -0.196] & -0.040 & [-0.074, -0.006] & [-0.109, 0.029] & -0.087 & [-0.123, -0.049] & [-0.159, -0.011]\\
korean & Previous best & $\rho$ & -0.016 & [-0.031, -0.003] & [-0.047, 0.011] & -0.013 & [-0.022, -0.005] & [-0.034, 0.003] & -0.005 & [-0.014, 0.003] & [-0.023, 0.011]\\
korean & Previous best & QWK & -0.128 & [-0.152, -0.105] & [-0.175, -0.082] & -0.208 & [-0.230, -0.187] & [-0.252, -0.167] & -0.144 & [-0.164, -0.125] & [-0.184, -0.107]\\
aesop & Previous best & $\rho$ & -0.023 & [-0.036, -0.011] & [-0.052, 0.000]$^{\dagger}$ & -0.022 & [-0.035, -0.011] & [-0.051, 0.000] & -0.037 & [-0.052, -0.024] & [-0.070, -0.012]\\
aesop & Previous best & QWK & -0.582 & [-0.600, -0.564] & [-0.617, -0.545] & -0.455 & [-0.479, -0.431] & [-0.501, -0.405] & -0.419 & [-0.443, -0.394] & [-0.466, -0.368]\\
koreanocean & Previous best & $\rho$ & -0.020 & [-0.039, -0.003] & [-0.061, 0.012] & -0.033 & [-0.056, -0.012] & [-0.084, 0.007] & -0.028 & [-0.049, -0.009] & [-0.074, 0.009]\\
koreanocean & Previous best & QWK & -0.000 & [-0.004, 0.004] & [-0.008, 0.008] & -0.110 & [-0.127, -0.093] & [-0.145, -0.077] & -0.090 & [-0.105, -0.076] & [-0.121, -0.062]\\
aesopocean & Previous best & $\rho$ & -0.000 & [-0.002, 0.001] & [-0.004, 0.003] & +0.006 & [-0.004, 0.016] & [-0.015, 0.028] & -0.018 & [-0.033, -0.005] & [-0.049, 0.009]\\
aesopocean & Previous best & QWK & -0.225 & [-0.246, -0.205] & [-0.266, -0.183] & +0.007 & [-0.013, 0.028] & [-0.033, 0.051] & -0.200 & [-0.222, -0.178] & [-0.244, -0.158]\\
koreanaesop & Previous best & $\rho$ & -0.005 & [-0.011, 0.002] & [-0.019, 0.008] & -0.003 & [-0.008, 0.001] & [-0.013, 0.006] & -0.000 & [-0.003, 0.003] & [-0.005, 0.006]\\
koreanaesop & Previous best & QWK & -0.343 & [-0.368, -0.318] & [-0.394, -0.294] & -0.292 & [-0.314, -0.270] & [-0.336, -0.249] & -0.243 & [-0.264, -0.223] & [-0.285, -0.204]\\
aesopkoreanocean & Previous best & $\rho$ & -0.003 & [-0.009, 0.002] & [-0.016, 0.008] & +0.002 & [-0.008, 0.013] & [-0.021, 0.024] & -0.001 & [-0.012, 0.009] & [-0.024, 0.020]\\
aesopkoreanocean & Previous best & QWK & -0.242 & [-0.266, -0.219] & [-0.289, -0.196] & -0.115 & [-0.132, -0.099] & [-0.150, -0.083] & -0.181 & [-0.197, -0.166] & [-0.214, -0.150]\\
\multicolumn{12}{l}{\textbf{3RB-A}}\\*
speechocean & Previous best & $\rho$ & -0.028 & [-0.047, -0.010] & [-0.070, 0.007] & -0.041 & [-0.064, -0.021] & [-0.091, -0.002] & -0.041 & [-0.063, -0.020] & [-0.091, -0.002]\\
speechocean & Previous best & QWK & -0.275 & [-0.314, -0.236] & [-0.350, -0.196] & -0.051 & [-0.085, -0.017] & [-0.119, 0.019] & -0.093 & [-0.130, -0.056] & [-0.165, -0.017]\\
korean & Previous best & $\rho$ & -0.019 & [-0.034, -0.005] & [-0.052, 0.009] & -0.019 & [-0.029, -0.010] & [-0.042, -0.002] & -0.013 & [-0.021, -0.005] & [-0.031, 0.002]\\
korean & Previous best & QWK & -0.163 & [-0.188, -0.137] & [-0.213, -0.113] & -0.241 & [-0.262, -0.220] & [-0.284, -0.201] & -0.179 & [-0.198, -0.161] & [-0.217, -0.144]\\
aesop & Previous best & $\rho$ & -0.010 & [-0.020, -0.000] & [-0.031, 0.010] & -0.019 & [-0.031, -0.008] & [-0.046, 0.001] & -0.020 & [-0.030, -0.011] & [-0.043, -0.002]\\
aesop & Previous best & QWK & -0.578 & [-0.596, -0.559] & [-0.615, -0.539] & -0.430 & [-0.453, -0.405] & [-0.475, -0.381] & -0.404 & [-0.427, -0.380] & [-0.450, -0.354]\\
koreanocean & Previous best & $\rho$ & -0.031 & [-0.050, -0.014] & [-0.072, 0.002] & -0.030 & [-0.050, -0.013] & [-0.074, 0.003] & -0.029 & [-0.048, -0.012] & [-0.070, 0.003]\\
koreanocean & Previous best & QWK & -0.019 & [-0.030, -0.007] & [-0.042, 0.004] & -0.044 & [-0.053, -0.036] & [-0.062, -0.028] & -0.046 & [-0.054, -0.038] & [-0.063, -0.030]\\
aesopocean & Previous best & $\rho$ & -0.001 & [-0.005, 0.003] & [-0.009, 0.007] & +0.004 & [-0.007, 0.014] & [-0.019, 0.026] & -0.002 & [-0.012, 0.009] & [-0.023, 0.021]\\
aesopocean & Previous best & QWK & -0.209 & [-0.231, -0.187] & [-0.254, -0.164] & -0.002 & [-0.021, 0.019] & [-0.040, 0.042] & -0.071 & [-0.092, -0.049] & [-0.114, -0.028]\\
koreanaesop & Previous best & $\rho$ & -0.004 & [-0.011, 0.003] & [-0.020, 0.010] & +0.001 & [-0.003, 0.005] & [-0.007, 0.009] & +0.006 & [0.002, 0.010] & [-0.001, 0.015]\\
koreanaesop & Previous best & QWK & -0.346 & [-0.371, -0.321] & [-0.396, -0.296] & -0.255 & [-0.278, -0.233] & [-0.301, -0.212] & -0.228 & [-0.250, -0.207] & [-0.271, -0.187]\\
aesopkoreanocean & Previous best & $\rho$ & -0.009 & [-0.016, -0.001] & [-0.026, 0.005] & -0.000 & [-0.011, 0.010] & [-0.023, 0.021] & -0.005 & [-0.015, 0.005] & [-0.027, 0.015]\\
aesopkoreanocean & Previous best & QWK & -0.243 & [-0.269, -0.219] & [-0.295, -0.195] & -0.167 & [-0.182, -0.152] & [-0.198, -0.138] & -0.169 & [-0.184, -0.155] & [-0.199, -0.142]\\
\end{longtable}
\noindent\textit{Note.} Positive differences favor the system; Human denotes human--other-rater consensus. Adjusted intervals cover 510 contrasts. Decisions use unrounded endpoints, not displayed zeros. QWK uses clipping and half-up rounding; Spearman retains raw scores. Previous accuracy leaders: 1G-S-aesopocean for correlation and 1G-S-koreanocean for QWK; fluency/prosody leaders: 5G-AMAE-koreanaesop for correlation and 5G-AMAE-koreanocean for QWK. A dagger marks seed-dependent interval exclusion.
\endgroup
\end{landscape}

\begin{landscape}
\begingroup\small\setlength{\tabcolsep}{4pt}\setlength{\LTcapwidth}{\linewidth}
\subsection{Score distributions}
The ridge distributions below use the Spearman-selected (1G-S) condition; Azure is independent of selection rule. Means, sample standard deviations, bias, and MAE use unrounded predictions on the 0--100 scale against the three-rater mean; no recalibration is fitted.
\begin{longtable}{@{}lrrrrrrrrrrrr@{}}
\caption{Score distributions for the 1G-S condition and Azure.}\\
\toprule
System & \multicolumn{4}{c}{Accuracy} & \multicolumn{4}{c}{Fluency} & \multicolumn{4}{c}{Prosody}\\
 & Mean & SD & Bias & MAE & Mean & SD & Bias & MAE & Mean & SD & Bias & MAE\\\midrule
\endfirsthead
\toprule
System & \multicolumn{4}{c}{Accuracy} & \multicolumn{4}{c}{Fluency} & \multicolumn{4}{c}{Prosody}\\
 & Mean & SD & Bias & MAE & Mean & SD & Bias & MAE & Mean & SD & Bias & MAE\\\midrule
\endhead
\bottomrule\endfoot
azure & 89.29 & 7.50 & 23.63 & 23.63 & 94.18 & 5.87 & 27.88 & 27.88 & 77.13 & 6.15 & 10.57 & 11.04\\
speechocean & 73.57 & 6.14 & 7.90 & 8.96 & 64.67 & 7.37 & -1.63 & 5.26 & 60.79 & 6.81 & -5.76 & 6.70\\
korean & 58.57 & 12.39 & -7.10 & 7.66 & 53.31 & 10.75 & -12.99 & 12.99 & 53.31 & 10.75 & -13.25 & 13.25\\
aesop & 38.49 & 13.31 & -27.17 & 27.17 & 44.43 & 10.97 & -21.87 & 21.87 & 44.14 & 10.62 & -22.42 & 22.42\\
koreanocean & 64.85 & 9.77 & -0.81 & 4.38 & 57.68 & 8.45 & -8.62 & 8.74 & 56.89 & 8.32 & -9.66 & 9.72\\
aesopocean & 55.27 & 11.66 & -10.40 & 10.51 & 56.63 & 10.99 & -9.67 & 9.79 & 55.25 & 9.89 & -11.31 & 11.33\\
koreanaesop & 50.09 & 14.23 & -15.58 & 15.59 & 51.09 & 12.44 & -15.21 & 15.21 & 50.26 & 12.36 & -16.29 & 16.29\\
aesopkoreanocean & 54.37 & 12.52 & -11.30 & 11.38 & 54.18 & 10.50 & -12.12 & 12.13 & 53.05 & 10.42 & -13.50 & 13.50\\
\end{longtable}
\section{Public-development comparisons}
The matched seven-family experiment uses fixed representation combinations and Spearman-based penalty selection (1G-S) over included-corpus development sets. All entries are development Spearman correlations, not final-test results. SpeechOcean uses the matched fitting procedure, not the original earlier deployment. Korean fluency and prosody share a source target. SO denotes SpeechOcean ($n=500$), Korean has $n=636$, and J-AESOP has $n=36$ development recordings; the small J-AESOP sample limits precision.
\begin{longtable}{@{}lrrrrrrrrr@{}}
\caption{Public-development Spearman correlations for the 1G-S condition.}\\
\toprule
System & \multicolumn{3}{c}{Accuracy} & \multicolumn{3}{c}{Fluency} & \multicolumn{3}{c}{Prosody}\\
 & SO & Korean & J-AESOP & SO & Korean & J-AESOP & SO & Korean & J-AESOP\\\midrule
\endfirsthead
\toprule
System & \multicolumn{3}{c}{Accuracy} & \multicolumn{3}{c}{Fluency} & \multicolumn{3}{c}{Prosody}\\
 & SO & Korean & J-AESOP & SO & Korean & J-AESOP & SO & Korean & J-AESOP\\\midrule
\endhead
\bottomrule\endfoot
speechocean & 0.816 & 0.699 & 0.967 & 0.886 & 0.668 & 0.856 & 0.890 & 0.653 & 0.929\\
korean & 0.735 & 0.759 & 0.969 & 0.837 & 0.778 & 0.919 & 0.829 & 0.778 & 0.938\\
aesop & 0.659 & 0.673 & 0.987 & 0.682 & 0.586 & 0.966 & 0.685 & 0.544 & 0.975\\
koreanocean & 0.808 & 0.755 & 0.964 & 0.876 & 0.770 & 0.924 & 0.876 & 0.771 & 0.949\\
aesopocean & 0.807 & 0.715 & 0.970 & 0.881 & 0.728 & 0.952 & 0.878 & 0.747 & 0.971\\
koreanaesop & 0.683 & 0.755 & 0.976 & 0.834 & 0.779 & 0.959 & 0.830 & 0.777 & 0.973\\
aesopkoreanocean & 0.804 & 0.748 & 0.971 & 0.875 & 0.769 & 0.953 & 0.872 & 0.770 & 0.974\\
\end{longtable}
\endgroup
\end{landscape}
\section{Korean supervision and representation sensitivity}
\subsection{Korean scoring and target mapping}
Han, Chung, and Kim \cite{han} define two 1--5 ratings: pronunciation accuracy and prosodic fluency. Accuracy addresses segmental clarity and errors; prosodic fluency combines stress, rhythm, intonation, speed, and pausing. The stored fields are \texttt{articulationScore} and \texttt{prosodyScore}. The second field supplies both fluency and prosody targets in the present implementation. Mapping $2.5(y-1)$ produces 0, 2.5, 5, 7.5, and 10 from the five integer levels; mapping does not equate the rubric with SpeechOcean or create an independent fluency annotation. Korean-only heads use the same features and targets for both constructs, yielding identical predictions in the evaluated configuration.
\begin{center}\begin{tabular}{@{}crp{5.4cm}p{5.4cm}@{}}\toprule
Source & Mapped & Accuracy (paraphrased) & Prosodic fluency (paraphrased)\\\midrule
1 & 0 & Extensive segmental difficulty; understanding is hard. & Strongly disrupted prosody, very slow delivery, excessive pauses.\\
2 & 2.5 & Frequent errors substantially burden understanding. & Awkward prosody and slow delivery with prolonged, poorly placed pauses.\\
3 & 5 & Recurring errors interfere with understanding. & Uneven prosody and rate, with frequent or misplaced breaks.\\
4 & 7.5 & Minor errors with little intelligibility impact. & Mostly steady delivery with mild prosodic or pausing irregularities.\\
5 & 10 & Clear segments without noticeable errors. & Natural prosody, rate, and phrase boundaries.\\\bottomrule
\end{tabular}\end{center}
\subsection{Ablation protocol}
The protocol was recorded before fitting on October 18, 2026 (repository: \nolinkurl{results/ablation-mac-v1/PROTOCOL.md}). The completed public-development sweep crossed seven training families, three constructs, and all 15 nonempty subsets of four feature blocks, giving 315 configurations and 945 corpus-level comparisons. X denotes XEUS, R XLS-R, P PhoneticXeus embeddings, and I the 17 phone diagnostics; I still requires PhoneticXeus inference, so omitting P does not necessarily avoid that encoder. Existing speaker-disjoint partitions, corpus-balanced weights, training-only standardization, and block gains were preserved. Each configuration selected a penalty from $\{0.001,0.01,0.1,1,10,100\}$ using equal-weight mean Spearman across all three development corpora, breaking ties by lower macro MAE and then larger penalty. This differs from the earlier included-corpus tuning; no corpus in these tables is an untouched cross-corpus test. Private recordings and ratings were excluded from fitting and selection.

Each candidate was compared with the fixed-feature baseline refitted under the same sweep protocol (X+P+I for accuracy, X+R otherwise). Candidates within 0.01 of the numerical best were shortlisted. Promotion required macro Spearman gain at least 0.01, no corpus correlation loss exceeding 0.02, no corpus MAE increase exceeding 0.10 on the 0--10 scale, and a positive lower 95\% paired speaker-bootstrap bound (2,000 draws, seed 42). Fewer blocks were preferred where benefits were inconclusive. These screening gates do not constitute correction for all search comparisons. Deployed heads were not automatically replaced.

The decision table distinguishes numerical maxima from candidates passing the complete gate. Two AESOP-only alternatives passed: X+P for fluency and X+P+I for prosody. Their macro gains over the refitted AESOP baselines were 0.0341 (95\% interval [0.0118, 0.0619]) for fluency and 0.0389 ([0.0161, 0.0638]) for prosody. Both exceed the 0.01 within-family threshold. Neither became the overall construct selection, and neither was substituted into the external seven-family evaluation. The overall retained choices were koreanocean X+P+I accuracy and aesopkoreanocean X+R fluency/prosody. Thus retaining the overall configurations does not mean that no individual family improved. XEUS-only macro correlations for the selected families were 0.8407 for koreanocean accuracy and 0.8644/0.8680 for aesopkoreanocean fluency/prosody, compared with 0.8424 and 0.8660/0.8721 for the retained combinations. These small development differences do not establish that all additional blocks are necessary; no XEUS-only private evaluation or matched resource benchmark is reported. The retained choices reflect the recorded development rule, not independently established representation optimality. Matrices report all 315 macro correlations; the accompanying CSV preserves all 945 corpus-specific correlations and MAEs.
\clearpage
\noindent\textbf{D.2 Ablation decisions (continued)}
\begin{longtable}{@{}lllrll@{}}
\toprule Family & Construct & Numerical best & Best $\rho$ & Retained/gated & Promotion\\\midrule\endhead
speechocean & accuracy & X+P+I & 0.8276 & X+P+I & No \\
speechocean & fluency & X+P+I & 0.8655 & X+R & No \\
speechocean & prosody & X+I & 0.8640 & X+R & No \\
korean & accuracy & X+R+I & 0.8322 & X+P+I & No \\
korean & fluency & X+I & 0.8544 & X+R & No \\
korean & prosody & X & 0.8560 & X+R & No \\
aesop & accuracy & X+P+I & 0.7813 & X+P+I & No \\
aesop & fluency & X+P & 0.7961 & X+P & Yes \\
aesop & prosody & X+P+I & 0.7893 & X+P+I & Yes \\
koreanocean & accuracy & X+P+I & 0.8424 & X+P+I & No \\
koreanocean & fluency & X+P & 0.8685 & X+R & No \\
koreanocean & prosody & X+P+I & 0.8691 & X+R & No \\
aesopocean & accuracy & X+P+I & 0.8309 & X+P+I & No \\
aesopocean & fluency & X+R+P+I & 0.8654 & X+R & No \\
aesopocean & prosody & X+R+P+I & 0.8653 & X+R & No \\
koreanaesop & accuracy & X+R & 0.8337 & X+P+I & No \\
koreanaesop & fluency & X+R+P & 0.8600 & X+R & No \\
koreanaesop & prosody & X+R & 0.8599 & X+R & No \\
aesopkoreanocean & accuracy & X+R+P+I & 0.8425 & X+P+I & No \\
aesopkoreanocean & fluency & X+R+P+I & 0.8710 & X+R & No \\
aesopkoreanocean & prosody & X+R & 0.8721 & X+R & No \\
\bottomrule\end{longtable}
\clearpage
\subsection{Representation matrices}
Columns S, K, A, SK, SA, KA, and SKA denote SpeechOcean, Korean, J-AESOP, and their respective combinations. Values are macro development Spearman correlations; X, R, P, and I denote XEUS, XLS-R, PhoneticXeus embeddings, and IPA diagnostics.
\begingroup\fontsize{8.5}{10}\selectfont\setlength{\tabcolsep}{5pt}\renewcommand{\arraystretch}{1.0}
\begin{longtable}{@{}lrrrrrrr@{}}
\caption{Representation ablation by construct and training family.}\\
\toprule Features & S & K & A & SK & SA & KA & SKA\\\midrule\endfirsthead
\toprule Features & S & K & A & SK & SA & KA & SKA\\\midrule\endhead
\bottomrule\endfoot
\multicolumn{8}{l}{\textbf{Accuracy}}\\*\midrule
X & 0.8208 & 0.8321 & 0.7655 & 0.8407 & 0.8226 & 0.8318 & 0.8336 \\
R & 0.6715 & 0.7460 & 0.3186 & 0.7646 & 0.7465 & 0.7766 & 0.7972 \\
P & 0.7860 & 0.7366 & 0.7292 & 0.8228 & 0.8065 & 0.7600 & 0.8273 \\
I & 0.6080 & 0.6742 & 0.6201 & 0.6961 & 0.3225 & 0.5164 & 0.4372 \\
X+R & 0.7957 & 0.8297 & 0.7414 & 0.8322 & 0.8257 & 0.8337 & 0.8361 \\
X+P & 0.8190 & 0.8030 & 0.7732 & 0.8350 & 0.8231 & 0.8159 & 0.8354 \\
X+I & 0.8253 & 0.8321 & 0.7522 & 0.8411 & 0.8249 & 0.8149 & 0.8362 \\
R+P & 0.7505 & 0.7690 & 0.6650 & 0.8192 & 0.8008 & 0.7871 & 0.8287 \\
R+I & 0.7290 & 0.7875 & 0.5625 & 0.7968 & 0.7765 & 0.7830 & 0.8085 \\
P+I & 0.7991 & 0.7750 & 0.7582 & 0.8308 & 0.8155 & 0.7171 & 0.8356 \\
X+R+P & 0.7962 & 0.8104 & 0.7453 & 0.8280 & 0.8235 & 0.8225 & 0.8350 \\
X+R+I & 0.7984 & 0.8322 & 0.7223 & 0.8363 & 0.8266 & 0.8183 & 0.8382 \\
X+P+I & 0.8276 & 0.8212 & 0.7813 & 0.8424 & 0.8309 & 0.8048 & 0.8409 \\
R+P+I & 0.7723 & 0.7932 & 0.7065 & 0.8260 & 0.8104 & 0.7740 & 0.8381 \\
X+R+P+I & 0.8048 & 0.8218 & 0.7565 & 0.8363 & 0.8304 & 0.8132 & 0.8425 \\
\midrule
\multicolumn{8}{l}{\textbf{Fluency}}\\*\midrule
X & 0.8551 & 0.8543 & 0.7782 & 0.8630 & 0.8588 & 0.8545 & 0.8644 \\
R & 0.7312 & 0.7734 & 0.4373 & 0.8213 & 0.8000 & 0.8100 & 0.8319 \\
P & 0.8428 & 0.7912 & 0.7640 & 0.8592 & 0.8495 & 0.8127 & 0.8599 \\
I & 0.6288 & 0.6757 & 0.6577 & 0.6876 & 0.4060 & 0.5606 & 0.5002 \\
X+R & 0.8364 & 0.8444 & 0.7620 & 0.8567 & 0.8599 & 0.8574 & 0.8660 \\
X+P & 0.8613 & 0.8465 & 0.7961 & 0.8685 & 0.8645 & 0.8578 & 0.8702 \\
X+I & 0.8569 & 0.8544 & 0.7671 & 0.8618 & 0.8576 & 0.8538 & 0.8662 \\
R+P & 0.8178 & 0.8162 & 0.7153 & 0.8555 & 0.8498 & 0.8415 & 0.8653 \\
R+I & 0.7833 & 0.8076 & 0.6342 & 0.8407 & 0.8291 & 0.8200 & 0.8458 \\
P+I & 0.8469 & 0.8099 & 0.7748 & 0.8570 & 0.8511 & 0.8004 & 0.8593 \\
X+R+P & 0.8481 & 0.8458 & 0.7730 & 0.8609 & 0.8654 & 0.8600 & 0.8696 \\
X+R+I & 0.8443 & 0.8516 & 0.7400 & 0.8593 & 0.8598 & 0.8533 & 0.8652 \\
X+P+I & 0.8655 & 0.8508 & 0.7866 & 0.8666 & 0.8644 & 0.8498 & 0.8677 \\
R+P+I & 0.8340 & 0.8283 & 0.7268 & 0.8576 & 0.8515 & 0.8333 & 0.8653 \\
X+R+P+I & 0.8509 & 0.8481 & 0.7635 & 0.8639 & 0.8654 & 0.8556 & 0.8710 \\
\midrule
\multicolumn{8}{l}{\textbf{Prosody}}\\*\midrule
X & 0.8631 & 0.8560 & 0.7817 & 0.8688 & 0.8599 & 0.8573 & 0.8680 \\
R & 0.7489 & 0.7740 & 0.3583 & 0.8172 & 0.7891 & 0.8178 & 0.8377 \\
P & 0.8313 & 0.7768 & 0.7520 & 0.8503 & 0.8416 & 0.7970 & 0.8526 \\
I & 0.6228 & 0.6834 & 0.6322 & 0.6960 & 0.4179 & 0.5379 & 0.5035 \\
X+R & 0.8410 & 0.8483 & 0.7504 & 0.8657 & 0.8653 & 0.8599 & 0.8721 \\
X+P & 0.8620 & 0.8384 & 0.7872 & 0.8664 & 0.8615 & 0.8508 & 0.8681 \\
X+I & 0.8640 & 0.8524 & 0.7627 & 0.8659 & 0.8609 & 0.8499 & 0.8693 \\
R+P & 0.8077 & 0.8070 & 0.6886 & 0.8517 & 0.8409 & 0.8296 & 0.8586 \\
R+I & 0.7789 & 0.8076 & 0.5893 & 0.8412 & 0.8229 & 0.8208 & 0.8498 \\
P+I & 0.8361 & 0.7979 & 0.7792 & 0.8527 & 0.8449 & 0.7793 & 0.8539 \\
X+R+P & 0.8444 & 0.8408 & 0.7602 & 0.8662 & 0.8647 & 0.8540 & 0.8707 \\
X+R+I & 0.8485 & 0.8512 & 0.7281 & 0.8663 & 0.8639 & 0.8495 & 0.8687 \\
X+P+I & 0.8635 & 0.8459 & 0.7893 & 0.8691 & 0.8649 & 0.8420 & 0.8690 \\
R+P+I & 0.8235 & 0.8210 & 0.7086 & 0.8522 & 0.8437 & 0.8214 & 0.8608 \\
X+R+P+I & 0.8485 & 0.8455 & 0.7554 & 0.8668 & 0.8653 & 0.8477 & 0.8715 \\
\midrule
\end{longtable}\endgroup
\section{Execution environment}
The same Windows/WSL machine used for the reported capacity test was inspected on October 19, 2026. The installation reports Ubuntu 26.04.1 LTS, kernel \texttt{6.18.54.1-microsoft-standard-WSL2}, Python 3.11.17, and PyTorch \texttt{2.11.0+cu128} (CUDA build 12.8). Hardware comprises an Intel Core i7-11700K with 16 logical CPUs and an NVIDIA RTX 3080 with 10 GiB VRAM; the host has 32 GB RAM, with approximately 24 GiB allocated to WSL and 8 GiB swap. The reported inference capacity test used the CPU, not CUDA. These versions document the inspected installation; the archived timing and memory trace, rather than a new benchmark, supplies the reported performance measurements.
\section{External rater backgrounds and procedure}
The original human-evaluation study reports three experienced university raters. The Korea-based rater had more than 20 years of teaching experience, including a decade of university pronunciation instruction and speaking assessment. The Japan-based rater had nearly 30 years of teaching experience and more than 10 years as a Cambridge English speaking assessor. The China-based rater had nearly 20 years of experience, including high-stakes speech rating and university speaking assessment in an international joint-degree program. All three held professorial positions, as confirmed by the study author.

Before evaluating the source recordings, the raters jointly assessed separate Stella recordings spanning proficiency levels and discussed discrepancies to establish shared criteria for segmental accuracy, fluency, and prosody. They then scored the 306 source recordings independently in randomized order, blinded to participant identity and automated scores. Ratings used a 0--100 scale in five-point increments, with communicative-effectiveness anchors labelled ``High Intelligibility'' above 85 and ``Low Intelligibility'' below 40. These are the documented general scale anchors, not a newly constructed analytic rubric for each dimension. The original study's authors served as its raters; their ratings were reused as the external reference for the present model evaluation. One recording was subsequently excluded for excessive duration, leaving the matched 305-recording cohort. Human calibration used separate recordings and did not fit or calibrate any automated scorer.

\end{document}